\documentclass[10pt,twocolumn,letterpaper]{article}

\usepackage{cvpr}
\usepackage{times}
\usepackage{epsfig}
\usepackage{graphicx}
\usepackage{amsmath}
\usepackage{amssymb}
\usepackage{subfig}
\usepackage{bm}
\usepackage{booktabs}
\usepackage[marginal]{footmisc}
\usepackage{diagbox}
\usepackage{enumitem}
\usepackage{algorithm}
\usepackage{multirow}
\usepackage{amsmath} 
\usepackage{xcolor}
\usepackage{authblk}
\usepackage[noend]{algpseudocode}
\usepackage[breaklinks=true,bookmarks=false]{hyperref}

\cvprfinalcopy % *** Uncomment this line for the final submission

\def\cvprPaperID{3645} % *** Enter the CVPR Paper ID here

\begin{document}

%%%%%%%%% TITLE
\title{OICSR: Out-In-Channel Sparsity Regularization for\\
Compact Deep Neural Networks}

\author[1,2,*]{Jiashi Li}
\author[1,2,*]{Qi Qi}
\author[1,2,\textdagger]{Jingyu Wang}
\author[1,2]{Ce Ge}
\author[1,2]{Yujian Li}
\author[1]{Zhangzhang Yue}
\author[1,2]{Haifeng Sun}
\affil[1]{State Key Laboratory of Networking and Switching Technology, Beijing University of Posts and
Telecommunications, Beijing 100876, P.R. China}
\affil[2]{EBUPT Information Technology Co., Ltd., Beijing 100191, P.R. China}
%{\tt\small \{lijiashi, qiqi, wangjingyu, gece, liyujian, yuezhangzhang, sunhaifeng\_1\}@ebupt.com}

\maketitle
{\renewcommand{\thefootnote}{\fnsymbol{footnote}}
\footnotetext[1]{Authors contributed equally}
\footnotetext[2]{Corresponding author: wangjingyu@bupt.edu.cn}
\footnotetext[3]{Please contact: lijiashi@bupt.edu.cn}}
 %将脚注符号设置为fnsymbol类型，即特殊符号表示
%%%%%%%%% ABSTRACT
\begin{abstract}
   Channel pruning can significantly accelerate and compress deep neural networks. Many channel pruning works utilize structured sparsity regularization to zero out all the weights in some channels and automatically obtain structure-sparse network in training stage. However, these methods apply structured sparsity regularization on each layer separately where the correlations between consecutive layers are omitted. In this paper, we first combine one out-channel in current layer and the corresponding in-channel in next layer as a regularization group, namely out-in-channel. Our proposed Out-In-Channel Sparsity Regularization (OICSR) considers correlations between successive layers to further retain predictive power of the compact network. Training with OICSR thoroughly transfers discriminative features into a fraction of out-in-channels. Correspondingly, OICSR measures channel importance based on statistics computed from two consecutive layers, not individual layer. Finally, a global greedy pruning algorithm is designed to remove redundant out-in-channels in an iterative way. Our method is comprehensively evaluated with various CNN architectures including CifarNet, AlexNet, ResNet, DenseNet and PreActSeNet on CIFAR-10, CIFAR-100 and ImageNet-1K datasets. Notably, on ImageNet-1K, we reduce 37.2\% FLOPs on ResNet-50 while outperforming the original model by 0.22\% top-1 accuracy. 
\end{abstract}
%%%%%%%%% BODY TEXT
\section{Introduction}
Convolutional neural networks (CNNs) have achieved significant successes in visual tasks, including image classification~\cite{he2016deep,krizhevsky2012imagenet,szegedy2015going}, object detection ~\cite{dai2016r,ren2015faster}, semantic segmentation~\cite{chen2015semantic,Long_2015_CVPR}, \etc. However, large CNNs suffer from massive computational and storage overhead. For instance, deep residual network ResNet-50~\cite{he2016deep} takes up about 190MB storage space, and needs more than 4 billion float point operations (FLOPs) to classify a single image. High demand for computation and storage resources severely hinders the deployment of large-scale CNNs in resource constrained devices such as mobile devices, wearable devices and Internet of Things (IoT) equipment.

Pruning~\cite{han2016deep,wen2016learning} is an important family of methods to slim neural network by removing redundant connections, channels and layers. Connection pruning gains high compression ratio but leads to non-structured sparsity of CNNs~\cite{wen2016learning}. The practical acceleration of non-structured sparsity is limited due to irregular memory access. Therefore, structured sparsity pruning~\cite{wen2016learning,Yu_2018_CVPR} becomes growing popular. 

Regularization-based channel pruning~\cite{liu2017learning,wen2016learning} is a popular direction of structured sparsity pruning. These works introduce structured sparsity regularization (structured regularization) into optimization objective of model training. Training with structured regularization transfers important features into a small quantity of channels and automatically obtains structure-sparse model. Pruning structure-sparse models keeps more features/accuracy compared with directly pruning non-sparse models~\cite{wen2016learning}. For channel-level pruning, existing regularization-based works apply structured regularization on each layer separately, and only enforce channel-level sparsity in out-channels. However, the corresponding in-channels in next layer are neglected and non-sparse. We call them as the separated structured regularization. Pruning one out-channel in current layer results in a dummy zero output feature map that in turn prunes a corresponding in-channel in next layer together. Without structure sparsity, useful features in in-channels of next layer are falsely discarded, which severely impair the representational capacity of the network.
\begin{figure*} [ht]
\centering
\subfloat [Consecutive fully-connected layers] {\includegraphics[width=0.36\textwidth]{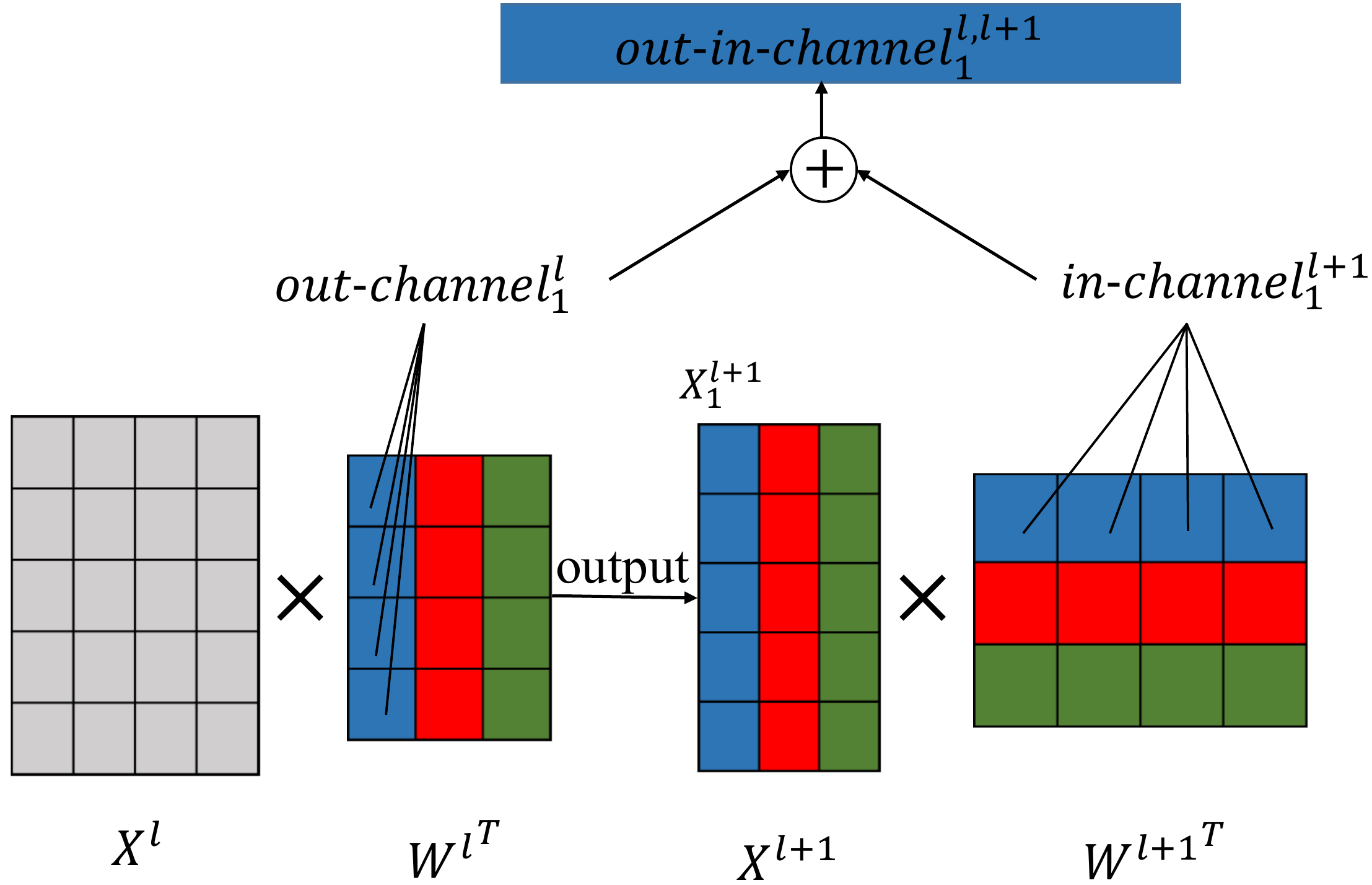}}
\quad\quad\quad\quad\quad	
\subfloat [Consecutive convolutional layers] {\includegraphics[width=0.45\textwidth]{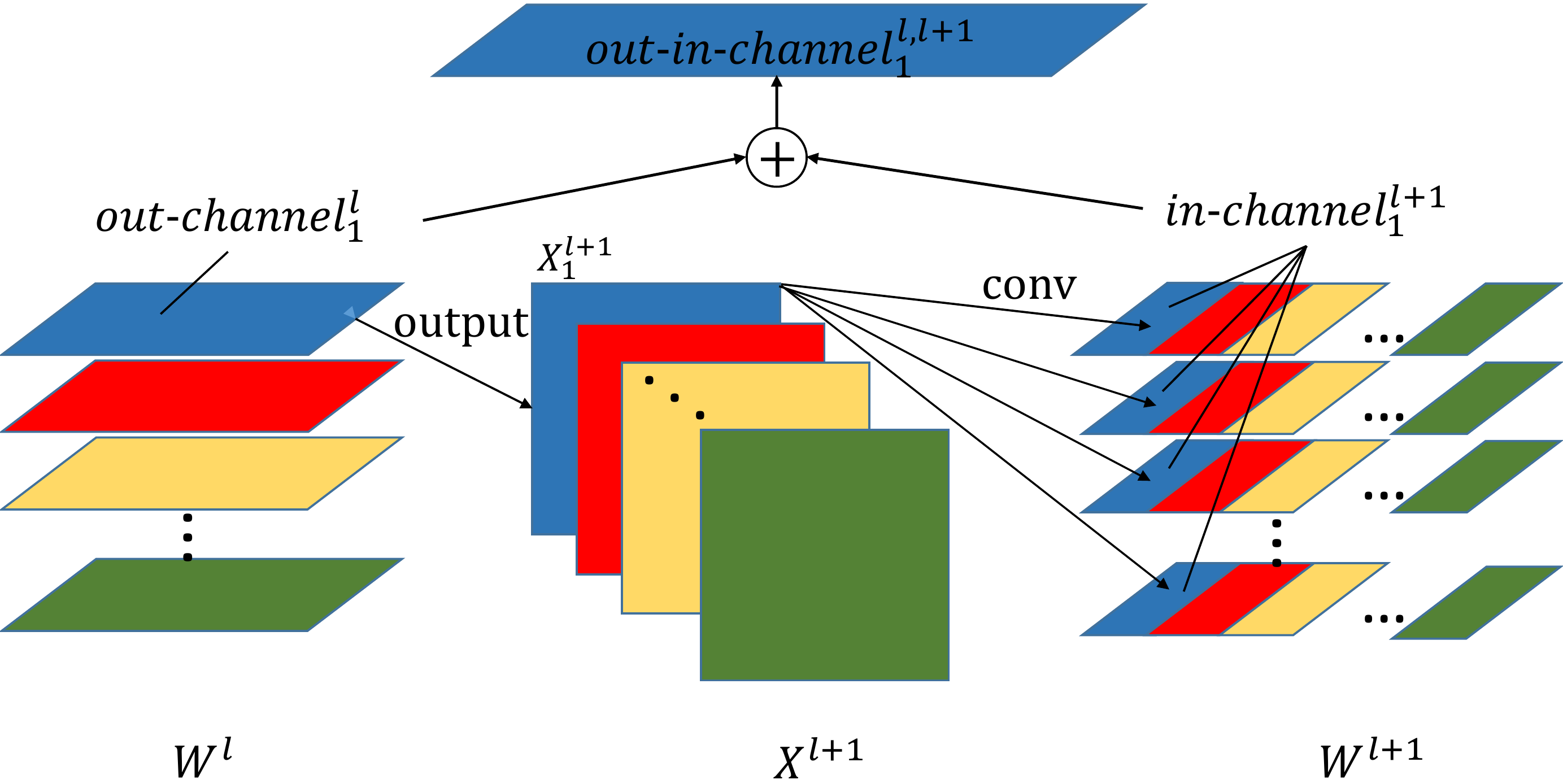}}
\caption{Correlations between two consecutive layers and the definition of out-in-channel. $out$-$channel^l_i$ is the weight vector of $i^{th}$ out-channel of $W^l$. $X^l$ is the input of $l^{th}$ layer and $X_i^{l+1}$ is the output of $X^l$ multiplied/convoluted by $out$-$channel^l_i$. Next, $X_i^{l+1}$ is multiplied/convoluted by $in$-$channel^{l+1}_i$ to obtain the input of next layer. The corresponding channels (marked by the same color) $out$-$channel^l_i$ and $in$-$channel^{l+1}_i$ of two consecutive layers tend to work cooperatively and are simultaneously pruned/saved. Therefore, $out$-$channel^l_i$ and $in$-$channel^{l+1}_i$ are regarded as one regularization group ($out$-$in$-$channel^{l,l+1}_i$) and are regularized together. Above '$\oplus$' denotes concatenation of $out$-$channel^l_i$ and $in$-$channel^{l+1}_i$.}
\label{out_in_channel}
\end{figure*}

In this paper, we propose a novel structured regularization form, namely Out-In-Channel Sparsity Regularization (OICSR), to learn more compact deep neural networks. Different from separated structured regularization, correlations between two consecutive layers are taken into account for channel pruning. An out-channel in current layer and the corresponding in-channel in next layer are combined to be a regularization group, namely out-in-channel. In training stage, features in one out-in-channel are simultaneously redistributed by OICSR. After training, features in redundant out-in-channels are thoroughly transferred to automatically selected important out-in-channels. Specially, the channel importance is measured based on statistics of two consecutive layers, not individual layer. To minimize accuracy loss induced by incorrect channel pruning, a greedy algorithm is proposed to globally prune redundant out-in-channels in an iterative way. As a result, pruning redundant out-in-channels induces negligible accuracy loss and it can be greatly compensated by the fine-tuning procedure. Our method achieves higher speedup ratio and compression compared with existing regularization-based methods. For ResNet-18~\cite{he2016deep} on CIFAR-10~\cite{krizhevsky2009learning} dataset, OICSR achieves 7.4$\times$ speedup and 11$\times$ parameters compression with tiny (0.19\%) top-1 accuracy drop.

The key advantages and major contributions of this paper can be summarized as follows:
\begin{itemize}[nosep]
   \item We propose a novel structured regularization form, namely OICSR, which takes account correlations between two consecutive layers to further retain predictive power of the compact network. 
   \item To minimize accuracy loss induced by incorrect channel pruning, OICSR measures channel importance based on statistics computed from two consecutive layers. A global greedy pruning algorithm is proposed to remove out-in-channels in an iterative way.
   \item To the best of our knowledge, this paper is the first attempt to present the evaluation of regularization-based channel pruning methods for very deep neural networks (ResNet-50 in this paper) on ImageNet~\cite{deng2009imagenet} dataset.
\end{itemize}

\section{Related Work}
Obtaining compact deep neural networks for speeding up inference and reducing storage overhead has been a long-studied project in both academia and industry. 

Recently, much attention has been focused on structured sparsity pruning to reduce network complexity. He \etal.~\cite{he2017channel} pruned channels by a LASSO regression based channel selection and least square reconstruction. Channel pruning was regarded as an optimization problem by Luo \etal~\cite{luo2017thinet:} and redundant channels were pruned by statistics of its next layer. Yu \etal~\cite{Yu_2018_CVPR} conducted feature ranking to obtain neuron/channel importance score and propagated it throughout the network. The neurons/channels with smaller importance scores were removed with negligible accuracy loss. Chin \etal~\cite{chin2018layer} considered channel pruning as a global ranking problem and compensated the layer-wise approximation error that improved the performance for various heuristic metrics. To reduce accuracy loss caused by incorrect channel pruning, redundant channels were pruned in a dynamic way in~\cite{he2018soft,lin2018accelerating}. Furthermore, Huang \etal.~\cite{huang2018learning} and  Huang \& Wang~\cite{huang2017data} trained pruning agents and removed redundant structure in a data-driven way. These methods directly pruned insignificant channels on non-structured sparse models, which may falsely abandon useful features and induce obvious accuracy decline.

More recent developments adopted structured regularization to learn structured sparsity in training stage. Zhang \etal.~\cite{zhou2016less} incorporated sparse constraints into objective function to decimate the number of channels in CNNs. Similarly, Wen \etal.~\cite{wen2016learning} utilized Group Lasso to automatically obtain channel, filter shape and layer level sparsity in CNNs during network training.  In~\cite{yoon2017combined}, group and exclusive sparsity regularization are combined to exploit both positive and negative correlations among features, while enforcing the network to be structure-sparse. Moreover, Liu \etal.~\cite{liu2017learning} proposed Network-Sliming which applied L1-norm on channel scaling factors. After training, channels with small-magnitude scaling factors are pruned. Zhang \etal.~\cite{zhang2018learning} adopted GrOWL regularization for simultaneous parameter sparsity and tying in CNNs learning. For channel pruning, these works automatically obtain channel-level sparse networks in training stage. Therefore, redundant channels are pruned with less accuracy decline. However, the above methods only apply separated structured regularization on out-channel in current layer but in-channels in next layer are neglected. Besides, these methods have not been accessed with very deep neural networks on ImageNet dataset. 
   
Low rank approximation~\cite{denton2014exploiting,Peng_2018_ECCV}, network quantization~\cite{courbariaux2016binarized,Jacob_2018_CVPR,rastegari2016xnor}, knowledge distillation~\cite{hinton2015distilling} and reinforcement learning~\cite{ashok2017n2n,he2018amc} are popular techniques to speedup and compress CNNs. These techniques can be combined with our channel pruning method for further improvement.

\section{Approach}
\subsection{Motivation}
We start by analyzing the drawbacks of separated structured regularization. The optimization objective of CNNs with separated structured regularization is formulized as:
\begin{equation}
J(W)=Loss(W, D)+\lambda R(W)+\lambda_{s}\sum_{l=1}^{L} R_{ss}(W^l)\label{equation 1}
\end{equation}
where $W$ is trainable weights across all the $L$ layers in CNNs and $D=\{(x_i, y_i)\}^N_{i=1}$ is a training dataset. $Loss(W, D)$ denotes the normal training loss on the dataset $D$. And $R(W)$ represents the non-structured regularization, \eg, L1 regularization and L2 regularization. The function $R_{ss}(\cdot)$ denotes the separated structured regularization applied on $L$ layers separately. $\lambda$ and $\lambda_{s}$ are the hyper-parameters of non-structured regularization and structured regularization. 

The correlations between two consecutive layers are illustrated in Fig.~\ref{out_in_channel}. Features in the $i^{th}$ out-channel of layer $l$ and the $i^{th}$ in-channel of layer $l+1$ are interdependent and tend to work cooperatively~\cite{luo2017thinet:,Yu_2018_CVPR}. Accordingly they should be regularized and redistributed together during training. However, $R_{ss}(W^l)$ regularizes layer $l$ separately. Suppose the $l^{th}$ layer is a fully connected layer with $W^l\in\mathbb{R}^{{OC}_l\times{IC}_l}$, where ${OC}_l$ and ${IC}_l$ are the dimensions of $W^l$ along the axes of out-channels and in-channels respectively. The separated structured regularization $R_{ss}(\cdot)$ applied on out-channels of $W^l$ for channel-level sparsity is:
\begin{equation}
R_{ss}(W^l)=\sum_{i=1}^{OC_l}\|W_{i,:}^l\|_{c}\label{equation 2}
\end{equation}
where $W_{i,:}^l$ is the weight vector of $i^{th}$ out-channel of $W^l$, $\sum_i\|\cdot\|_{c}$ is a specific structured regularization term which can effectively zero out all weights in some out-channels, such as Group Lasso~\cite{yuan2006model}, CGER~\cite{yoon2017combined} and GrOWL~\cite{figueiredo2016ordered}. The separated Group Lasso for channel-level sparsity is:
\begin{equation}
R_{ss}^*(W^l)=\sum_{i=1}^{OC_l}\sqrt{\sum_{j}({W_{i,j}^l})^2}\label{equation 3}
\end{equation}	
And the derivative of $R_{ss}^*(W^l)$ with respect to $W_{i,j}^l$ is:
\begin{equation}
\frac{\partial{R_{ss}^*(W^l)}}{\partial{W_{i,j}^l}}=\frac{W_{i,j}^l}{\sqrt{\sum_{j}({W_{i,j}^l})^2}}\label{gradient}
\end{equation}  
According to the gradient descent algorithm, the update of $W_{i,j}^l$ is influenced by all the weights in the $i^{th}$ out-channel of layer $l$.  And the decrease of $\sum_{j}({W_{i,j}^l})^2$ will boosts the decrease of $W_{i,j}^l$. The smaller value of $\sum_{j}({W_{i,j}^l})^2$ is, the faster $W_{i,j}^l$ decreases. Therefore, $R_{ss}^*(\cdot)$ can effectively zero out all the weights in some out-channels in training stage.

 The critical issue of separated structured regularization is that the correlations between two consecutive layers of CNNs are disregarded. It separately regularizes and enforces out-channels of each layer to be sparse. After training with separated structured regularization, features in the $i^{th}$ out-channel of layer $l$ may be squeezed to the $i^{th}$ in-channel of layer $l+1$, instead of the rest out-channels of layer $l$. Pruning the $i^{th}$ out-channel of layer $l$ results in pruning the $i^{th}$ in-channel of layer $l+1$ together. Important features in the $i^{th}$ in-channel of layer $l+1$ may be falsely discarded that losses massive accuracy. Moreover, the separated structured regularization fails to maximally prune redundant channels and utilize the representational capacity of CNNs.

\subsection{Out-In-Channel Sparsity Regularization}
We propose out-in-channel sparsity regularization to tackle the drawbacks of separated structured regularization. The definition of out-in-channel is demonstrated in Fig.~\ref{out_in_channel}. The corresponding channels of two consecutive layers work cooperatively and are simultaneously pruned/saved. Therefore, OICSR concatenates $out$-$channel^l_i$ with $in$-$channel^{l+1}_i$ as one regularization group $out$-$in$-$channel^{l,l+1}_i$. The optimization objective with OICSR of CNNs can be given as follows:
\begin{equation}
\begin{split}
&J(W)=Loss(W, D)+\lambda R(W) \\
&\qquad \qquad+\lambda_{s}\sum_{l=1}^{L-1}R_{oic}(W^l, W^{l+1})\label{equation 4}
\end{split}
\end{equation} 
where $R_{oic}(W^l, W^{l+1})$ is the out-in-channel sparisity regularization which regularizes out-in-channels of layer $l$ and layer $l+1$ together. OICSR of two consecutive fully-connected layers is given as:
\begin{equation}
R_{oic}(W^l, W^{l+1})=\sum_{i=1}^{OC_l}\|W_{i,:}^l\oplus W_{:,i}^{l+1}\|_{oic}\label{equation 5}
\end{equation}
where $\sum_i\|\cdot\|_{oic}$ is a specific structured regularization term in OICSR form which can simultaneously zero out all weights in some out-in-channels. The symbol $\oplus$ denotes concatenation of $W_{i,:}^l$ and $W_{:,i}^{l+1}$. For two consecutive convolutional layers, with $W^l\in\mathbb{R}^{{OC}_l\times{IC}_l\times{H_l}\times{W_l}}$ and $W^{l+1}$, where ${H}_l$ and ${W}_l$ denote the height and width respectively, we first reshape $W^l$ and $W^{l+1}$ to 2D matrices, \ie $W^l\in\mathbb{R}^{{OC}_l\times(IC_lH_lW_l)}$ and $W^{l+1}\in\mathbb{R}^{(OC_{l+1}H_{l+1}W_{l+1})\times{IC}_{l+1}}$. Then, OICSR of two consecutive convolutional layers can be similarly formulated as Eq.~\ref{equation 5}. 

The other structured regularization terms can be extended into OICSR form in a similar way. The derivative of $R_{oic}^*(W^l, W^{l+1})$ with respect to $W_{i,j}^l$ is:
\begin{equation}
\frac{\partial{R_{oic}^*(W^l, W^{l+1})}}{\partial{W_{i,j}^l}}=\frac{W_{i,j}^l}{\sqrt{\sum_{j}({W_{i,j}^l})^2+\sum_{j}({W_{j,i}^{l+1}})^2}}\label{gradient_oic}
\end{equation}   
By incorporating $R_{oic}^*(\cdot)$ into objective function, the update of $W_{i,j}^l$ is synchronously influenced by weights in the $i^{th}$ out-channel of layer $l$ and weights in the $i^{th}$ in-channel of layer $l+1$. The decrease of some weights will boost the decrease of the remaining weights in the same out-in-channel. In training stage, the out-in-channels with more redundant (nearly zero) weights will be gradully turned to insignificant out-in-channels by $R_{oic}^*(\cdot)$. Compared with $R_{ss}^*(\cdot)$ in Eq.~\ref{equation 3} and Eq.~\ref{gradient}, $R_{oic}^*(\cdot)$ can synchronously zero out all the weights in some less important out-in-channels and automatically obtain out-in-channel level sparse model in training stage.

OICSR regards one out-in-channel as a regularization group in which features are simultaneously redistributed during network training. After training, features in redundant out-in-channels are thoroughly transferred to important out-in-channels. As a result, OICSR is able to prune more redundant out-in-channels in large networks with less accuracy loss. Actually, the correlations between layer $l$ and layer $l+n\;(n\geqslant2)$ are too complex to be formulated in structured regularization form. It is a trade-off between practicability and effectiveness to consider the correlation of two consecutive layers.
\begin{figure} [tbp]
\centering
\includegraphics[width=0.48\textwidth]{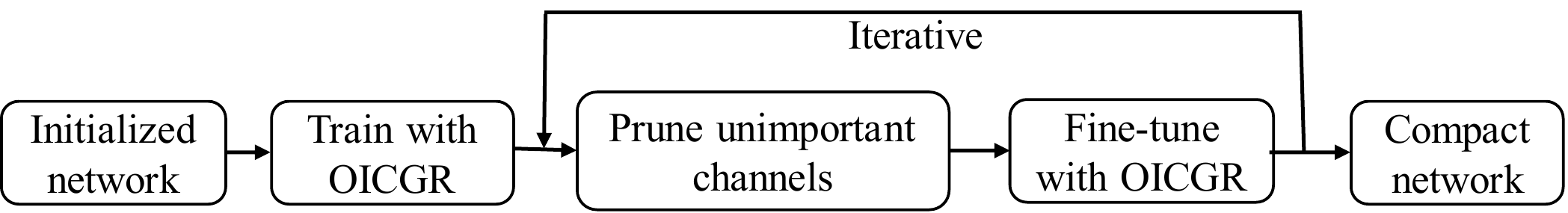}
\caption{Iterative channel pruning procedure with OICSR.}
\label{iterative}
\end{figure} 
\subsection{Criterion of Channel Importance}
For computational efficiency, the channel energy is chosen as the channel importance metric. Existing regularization-based methods~\cite{lebedev2016fast,liu2017learning,wen2016learning,yoon2017combined} only utilize statistics of individual layer to guide the channel pruning. In this paper, the channel importance of separated Group Lasso and non-structured regularization is defined as:
\begin{equation}
{E}_i^{l}=\|W_{i,:}^l\|_{2}^2=\sum_{j}({W_{i,j}^l})^2 \label{equation 7}
\end{equation}
where ${E}_i^{l}$ is the energy of the $i^{th}$ out-channel of layer $l$. The statistical imformation of next layer are abandoned that may cause incorrect selection of redundant channels. In particular, OICSR measures channel importance based on statistical imformation of two consecutive layers. The channel importance of  Group Lasso in OICSR form is given as:
\begin{equation}
{E}_i^{l, l+1}=\|W_{i,:}^l\oplus W_{:,i}^{l+1}\|_{2}^2=\sum_{j}({W_{i,j}^l})^2+\sum_{j}({W_{j,i}^{l+1}})^2\label{equation 8}
\end{equation}
where ${E}_i^{l, l+1}$ is the energy of the $i^{th}$ out-in-channel of layer $l$ and layer $l+1$. The higher the energy, the more important the out-in-channel is.  

\subsection{Channel Pruning Framework}
With initialized deep neural networks, our iterative channel pruning procedures are illustrated in Fig.~\ref{iterative}. In fact, it is puzzling to manually determine the redundancy and channel pruning ratio for each layer. Therefore, a global greedy pruning algorithm is proposed to minimize the accuracy loss caused by incorrect channel pruning. As shown in Algorithm~\ref{alg1}, in each iteration, redundant out-in-channels across all layers are globally selected and greedily removed until reaching the preset FLOPs pruning ratio. 

\begin{algorithm}
\caption{Global greedy pruning algorithm}
\label{alg1}
\begin{algorithmic}[1]
\Require Training dataset $\mathcal{D}$, initialzed model $\mathcal{W}$, number of pruning iteration $\mathcal{T}$, FLOPs pruning ratio $\mathcal{P}\in\mathbb{R}^{\mathcal{T}}$
\State $\mathcal{W}^{(0)}\leftarrow\ \texttt{train}(\mathcal{W}, \mathcal{D})$ with OICSR from scratch
\For {$\textit{t}=1$ to $\mathcal{T}$}
\State $\tilde{E}\leftarrow\varnothing$
\State // \textit{global channel selection}
\For {$\textit{l}=1$ to $L-1$}
\For {$\textit{i}=1$ to $OC_l$}
\State $\tilde{E}\leftarrow \tilde{E}\cup\{E^{l,l+1}_i\}$ as Eq.~\ref{equation 8}
\EndFor
\EndFor
\State $\tilde{E}=\texttt{sort}(\tilde{E})$
\State // \textit{greedy channel pruning}
\Repeat 
\State // \textit{remove the corresponding channel with $\tilde{E}(0)$}
\State $\mathcal{W}^{(t-1)}\leftarrow \texttt{prune}(\mathcal{W}^{(t-1)}, \tilde{E}(0))$
\State $\tilde{E} \leftarrow \tilde{E} \backslash \tilde{E}(0)$
\Until{$\texttt{flops}(\mathcal{W}^{(t-1)})} < (1-\mathcal{P}_t)\cdot{\texttt{flops}(\mathcal{W}^{(0)})}$
\State $\mathcal{W}^{(t)}\leftarrow \mathcal{W}^{(t-1)}$
\State $\mathcal{W}^{(t)}\leftarrow\ \texttt{fine-tune}(\mathcal{W}^{(t)}, \mathcal{D})$ with OICSR
\EndFor
\Ensure The compact model $\mathcal{W}^{(\mathcal{T})}$
\end{algorithmic}
\end{algorithm}
Compared with single pass pruning, iterative pruning leads to smoother pruning process with less accuracy drop. Pruning a whole layer is detrimental to the network~\cite{chin2018layer,liu2017learning}. Accordingly, we set a constraint that no more than 50\% of out-in-channels in two consecutive layers are pruned in one channel pruning iteration. 
\begin{figure*} [t]
\centering
\subfloat [CifarNet on CIFAR-10] {\includegraphics[width=0.24\textwidth]{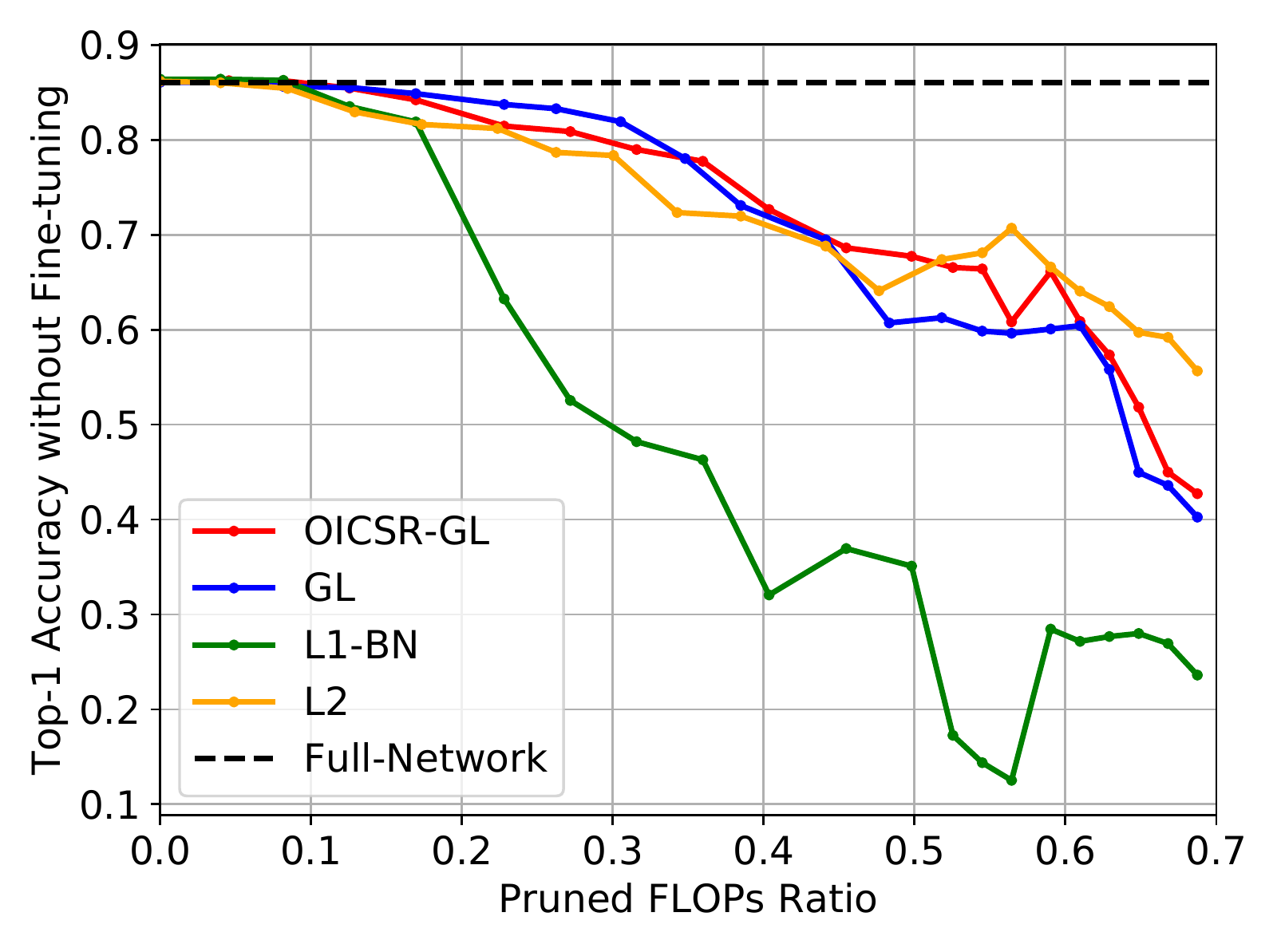}\label{cifarnet-cifar10}}
\hspace{1mm}
\subfloat [ResNet-18 on CIFAR-10] {\includegraphics[width=0.24\textwidth]{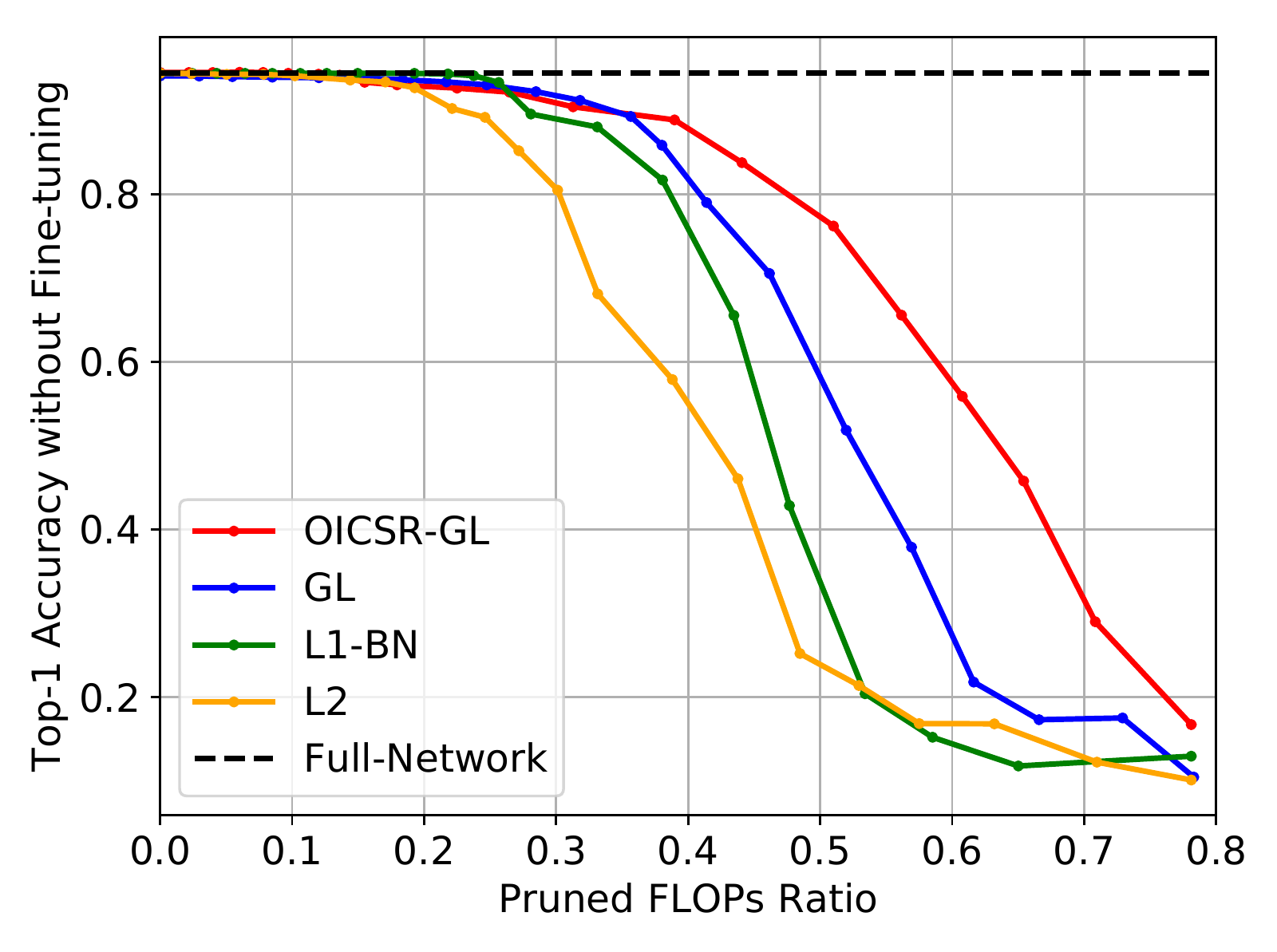}}
\hspace{1mm}
\subfloat [DenseNet-89 on CIFAR-10] {\includegraphics[width=0.24\textwidth]{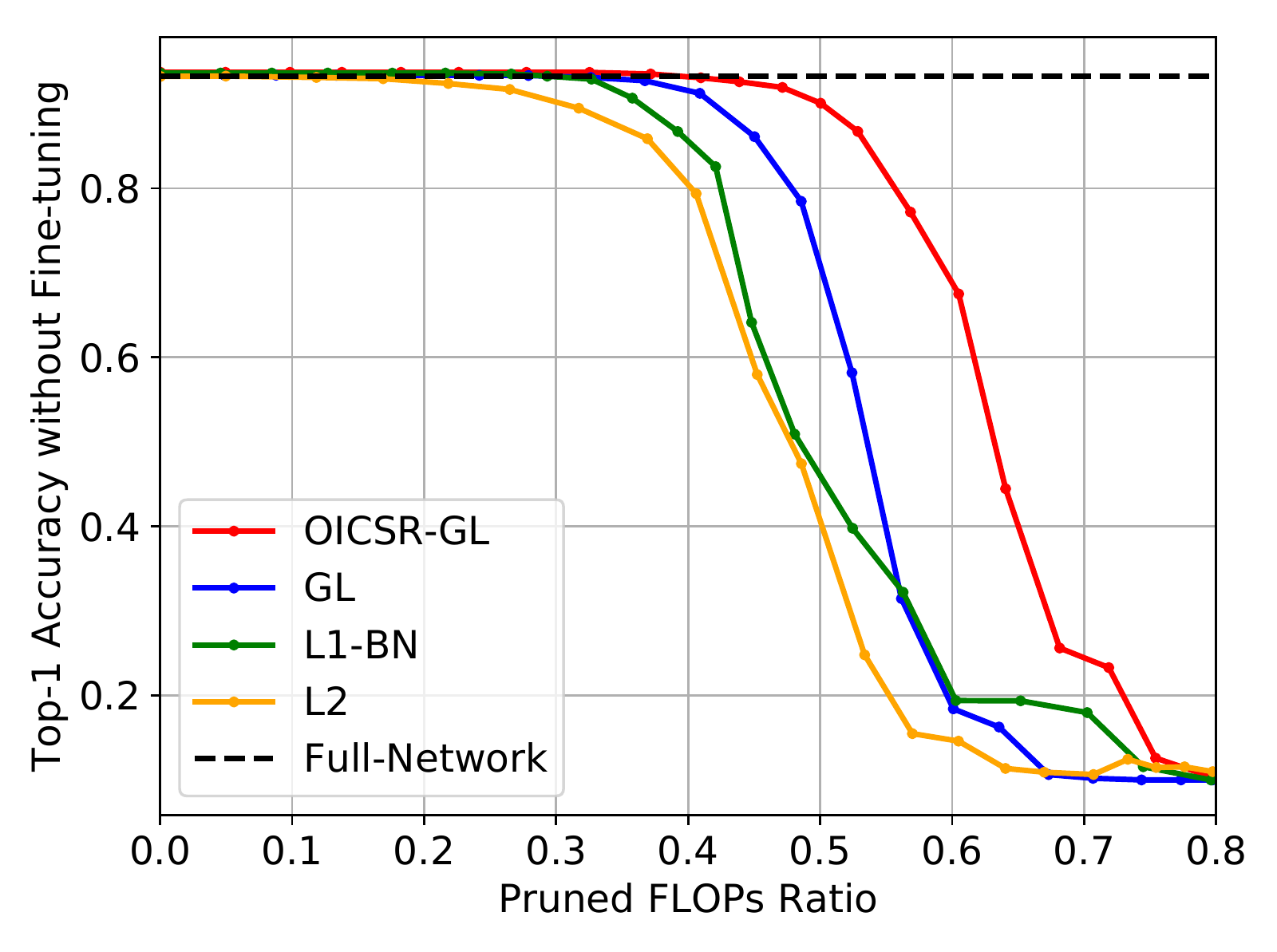}\label{DenseNet89-cifar10}}
\hspace{1mm}
\subfloat [AlexNet on ImageNet-1K] {\includegraphics[width=0.24\textwidth]{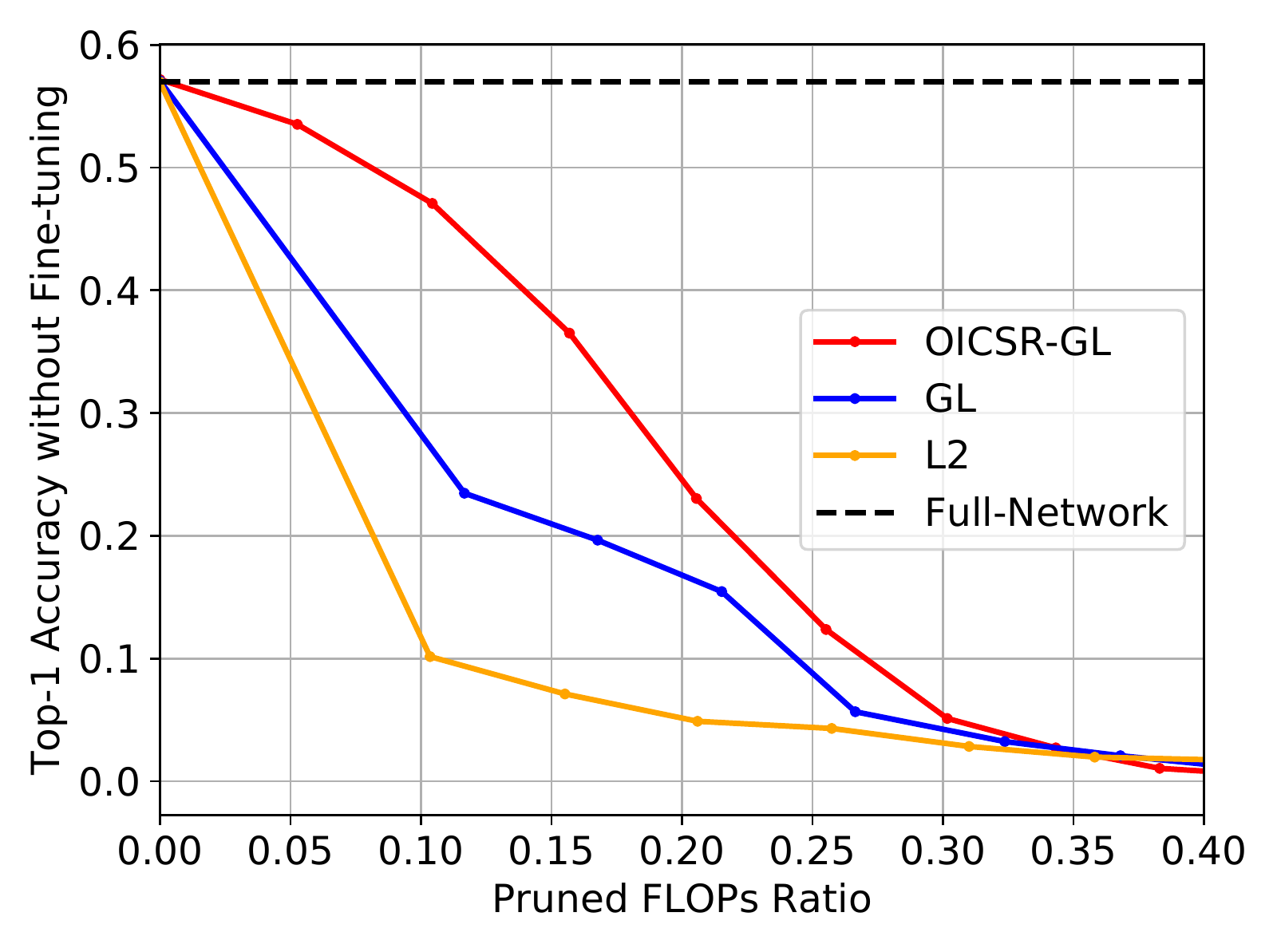}}
\hfil
\subfloat [CifarNet on CIFAR-100] {\includegraphics[width=0.24\textwidth]{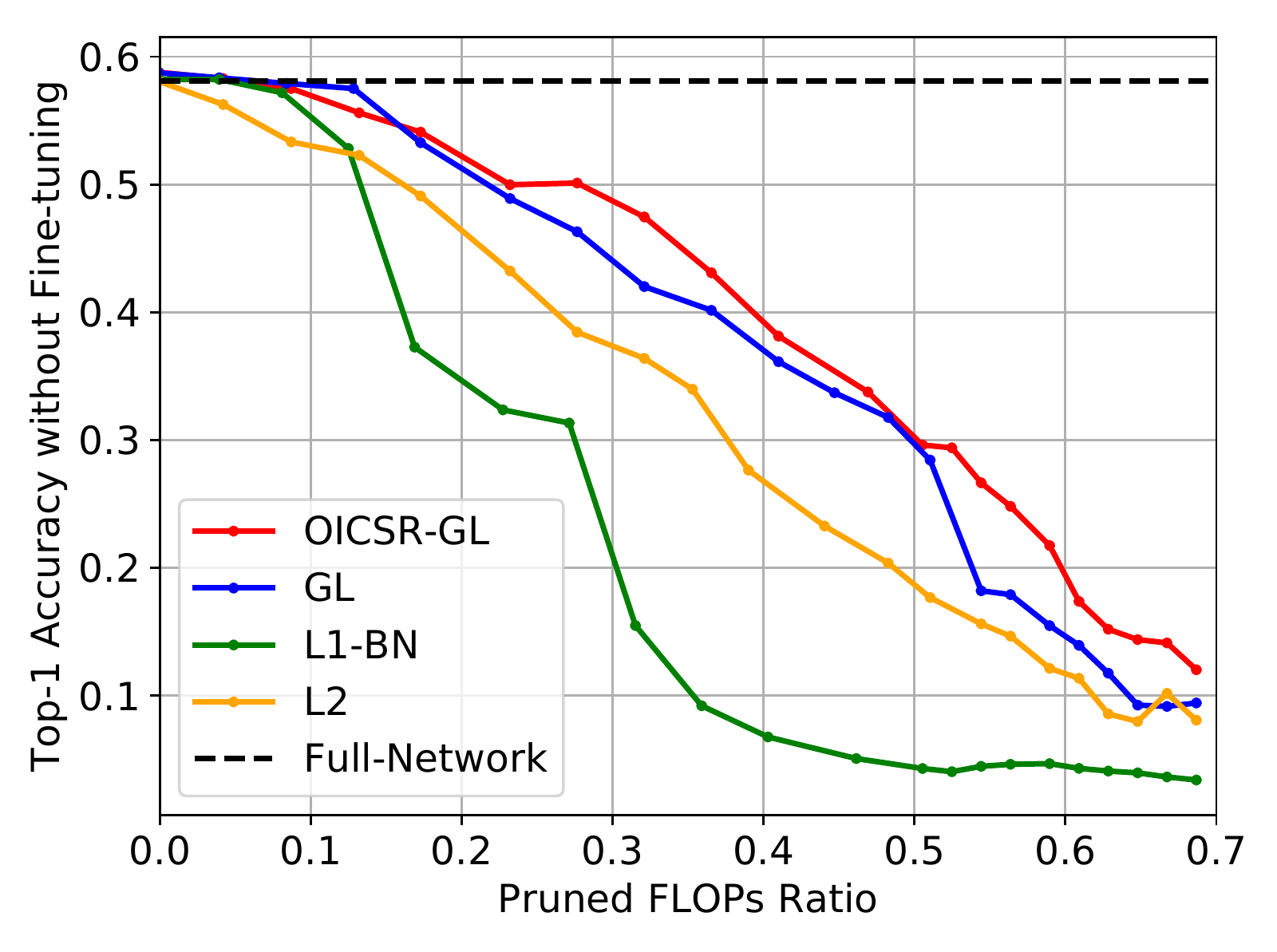}}
\hspace{1mm}
\subfloat [ResNet-56 on CIFAR-100] {\includegraphics[width=0.24\textwidth]{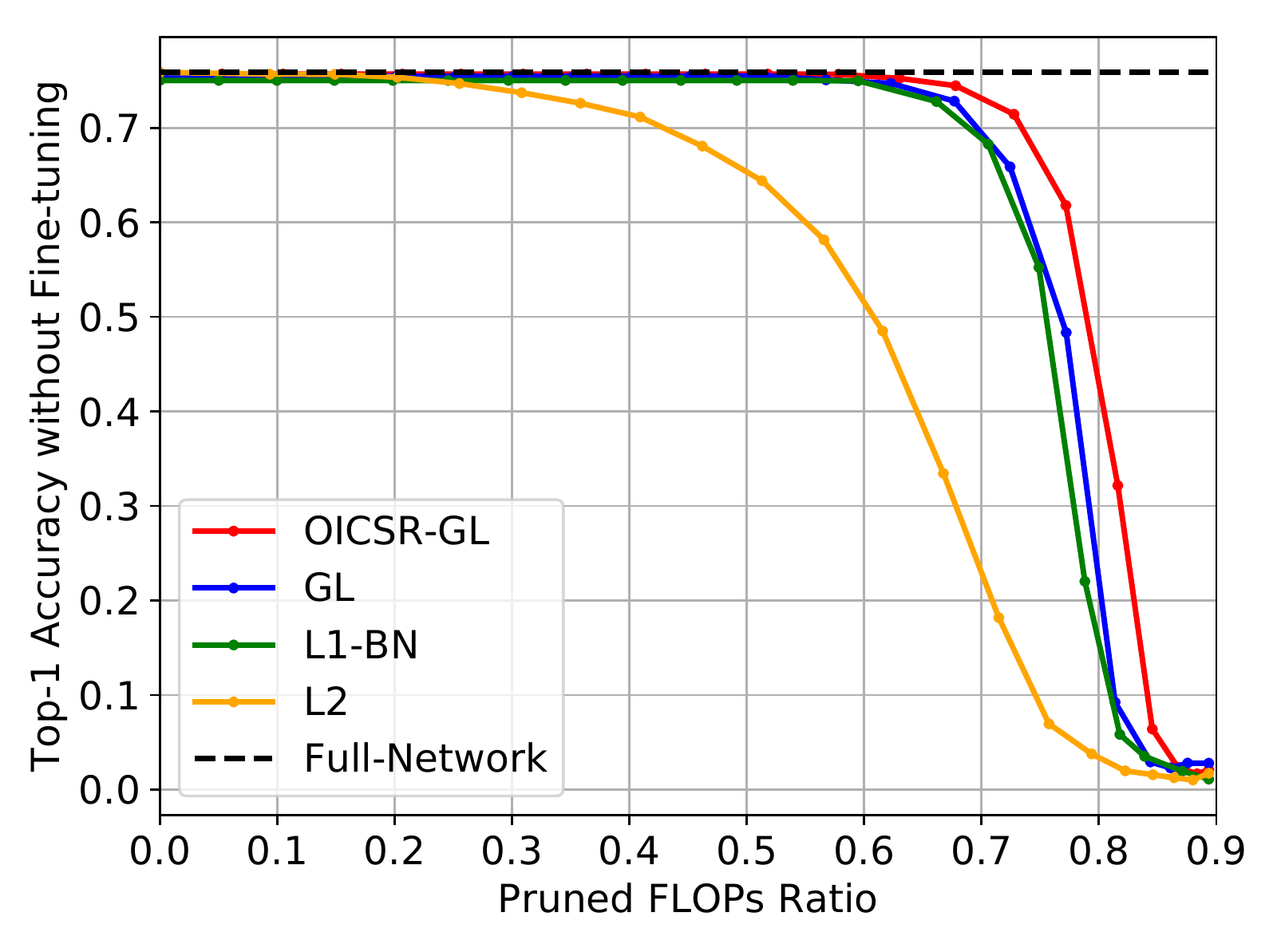}}
\hspace{1mm}
\subfloat [PreActSeNet-18 on CIFAR-100] {\includegraphics[width=0.24\textwidth]{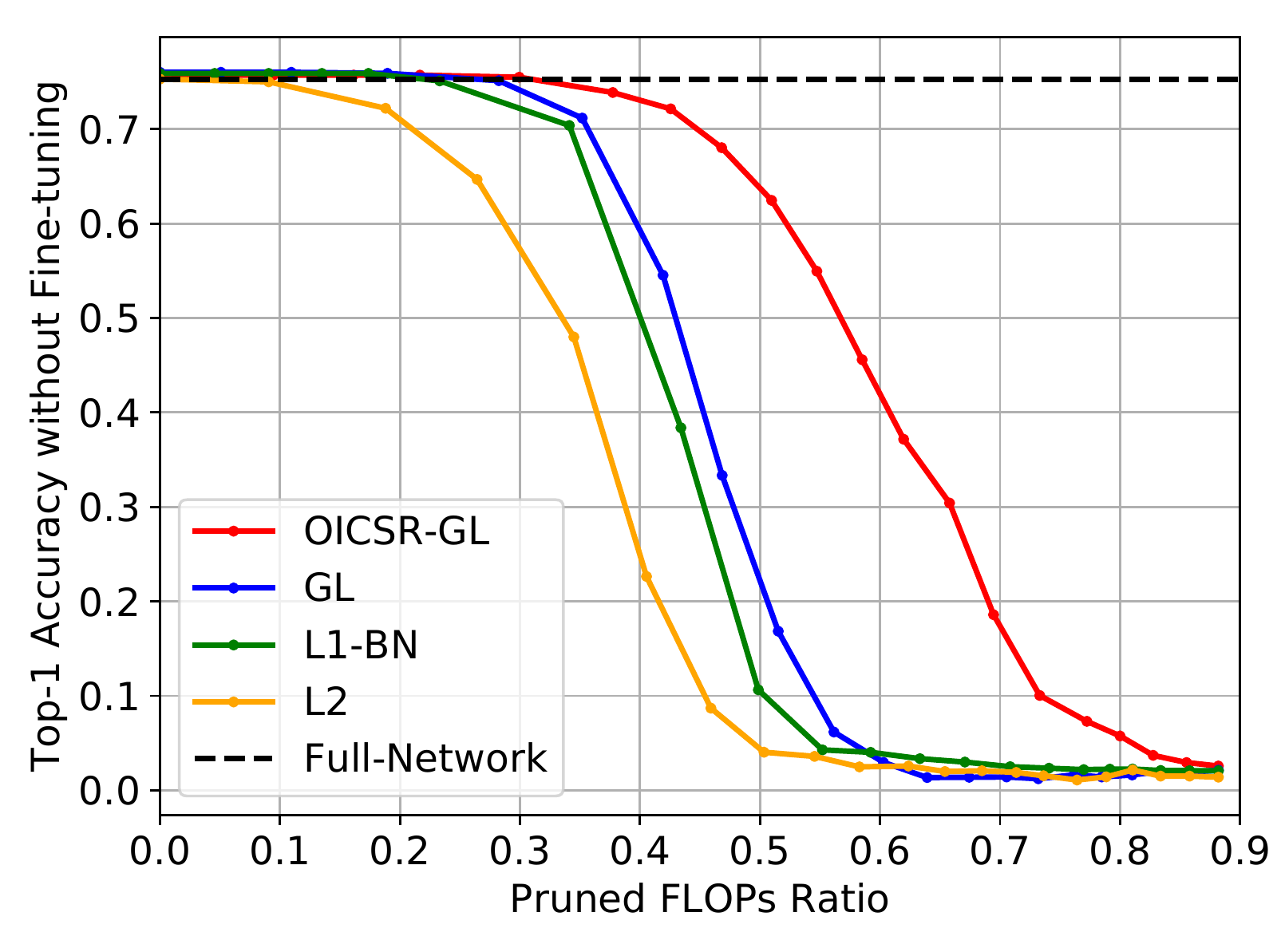}\label{PreActSeNet-cifar100}}
\hspace{1mm}
\subfloat [ResNet-50 on ImageNet-1K] {\includegraphics[width=0.24\textwidth]{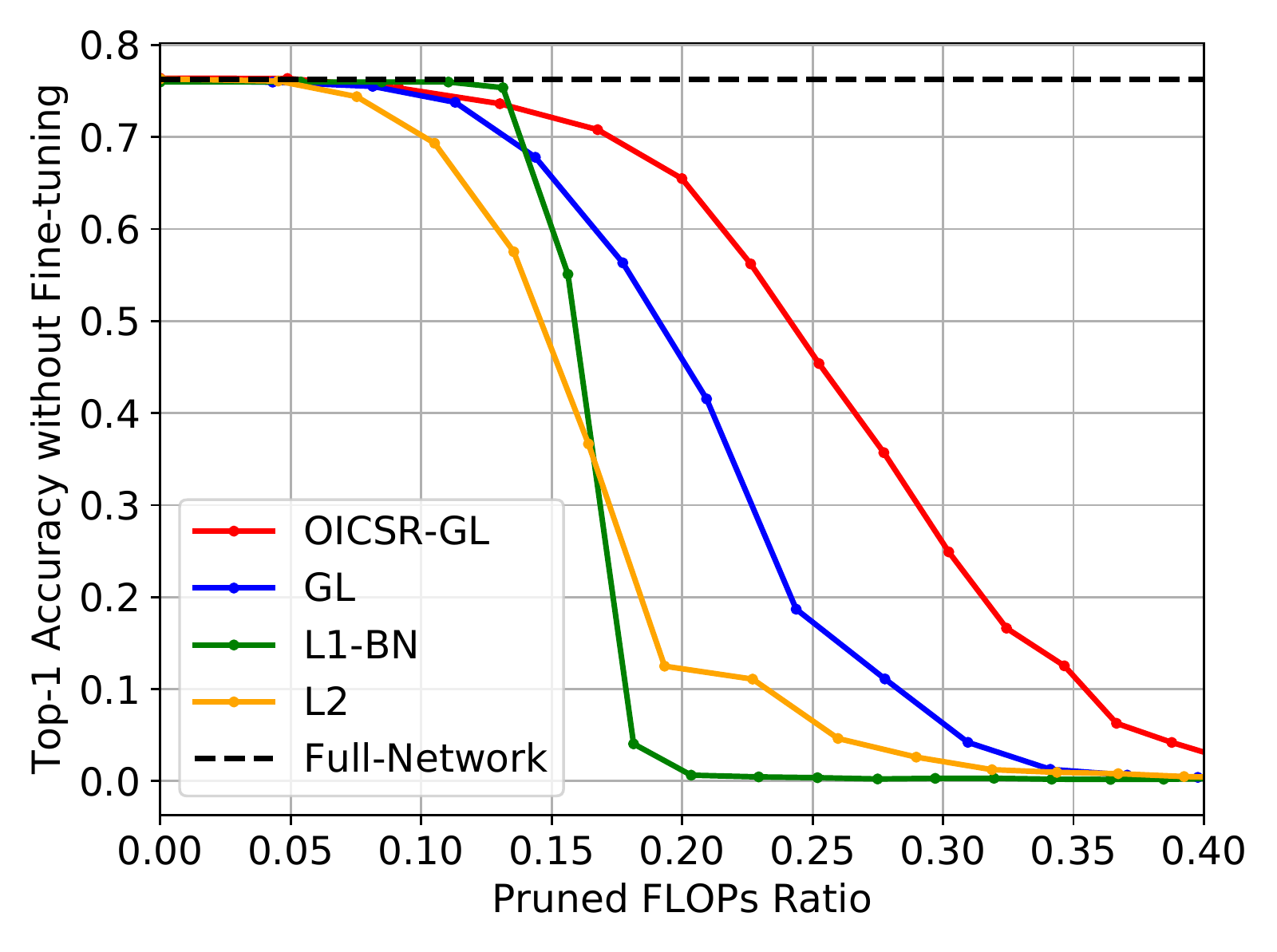}}
\caption{Comparison between OICSR-GL and baselines of the trade-off between top-1 accuracy (without fine-tuning) and pruned FLOPs ratio. L1-BN~\cite{liu2017learning} can not be appied on the variant AlexNet\textsuperscript{\ref{pytorch-vision}} which has no batch normalization layers. Obviously, OICSR-GL generally has less accuracy drop compared with baselines under the same pruned FLOPs.}
\label{accuracy-bf}
\end{figure*}

Fine-tuning is an important process after channel pruning. To the best of our knowledge, we are the first to fine-tune the pruned network with structured regularization. Fine-tuning with OICSR simultaneously recovers the diminished accuracy of channel pruning in last step and enforces channel-level sparsity on the pruned model. Therefore, the next iteration of channel pruning is smoothly conducted after fine-tuning.

\section{Experiments} \label{experiment}
In this section, we evaluate the effectiveness of OICSR on CIFAR-10~\cite{krizhevsky2009learning}, CIFAR-100~\cite{krizhevsky2009learning}, ImageNet-1K~\cite{deng2009imagenet} datasets using popular CNNs architectures: CifarNet~\cite{krizhevsky2009learning}, AlexNet~\cite{krizhevsky2012imagenet}, ResNet~\cite{he2016deep}, DenseNet~\cite{huang2017densely} and SeNet~\cite{hu2017squeeze}. OICSR is mainly compared with non-structured regularization and separated structured regularization to demonstrate its superiority. Moreover, we also compare OICSR with other state-of-the-art channel pruning methods~\cite{chin2018layer,he2018soft,he2017channel,huang2018learning,huang2017data,lin2018accelerating,luo2017thinet:,molchanov2017pruning,Yu_2018_CVPR}. All the experiments are implemented using PyTorch~\cite{paszke2017automatic} on four NVIDIA P100 GPUs.
\subsection{Experimental Setting}
For CIFAR-10/100 datasets, OICSR is evaluated with CifarNet\footnote{https://github.com/tensorflow/models/blob/master/research/slim/nets\label{CifarNet}}, ResNet-18\footnote{https://github.com/kuangliu/pytorch-cifar/tree/master/models \label{pytorch-cifar}}, ResNet-56\textsuperscript{\ref{pytorch-cifar}}, DenseNet-89\textsuperscript{\ref{pytorch-cifar}} and PreActSeNet-18\textsuperscript{\ref{pytorch-cifar}}. OICSR is also accessed with AlexNet\footnote{https://github.com/pytorch/vision/tree/master/torchvision/models \label{pytorch-vision}} and ResNet-50\textsuperscript{\ref{pytorch-vision}} on ImageNet-1K dataset. All the initialized networks are trained from scratch using SGD optimizer with a weight decay $10^{-4}$ and Nesterov momentum~\cite{sutskever2013importance} of 0.9. On CIFAR-10/100 datasets, we train networks using mini-batch size 100 for 160 epochs. On ImageNet-1K dataset, we train AlexNet and ResNet-50 with mini-batch size 256 for 90 and 120 epochs, respectively. All the accuracies on ImageNet-1K dataset are tested on the validation dataset using the single view center crop.

The hyper-parameter $\lambda_s$ balances the normal training loss and the structured sparsity. We empirically recommend choosing relatively large $\lambda_s$ for simple task but small $\lambda_s$ for complex task. The hyper-parameter $\lambda_s$ is set to $10^{-4}$ for networks (except for DenseNet-89 with $5\times10^{-5}$) on CIFAR-10/100 dataset and $10^{-5}$ for ImageNet-1K dataset.

Considering that fully-connected layers are much important in CifarNet and AlexNet and the majority of FLOPs is contributed by convolutional layers, only convolutional layers in these networks are regularized and pruned. 
\begin{figure} [t]
\centering
\includegraphics[width=0.49\textwidth]{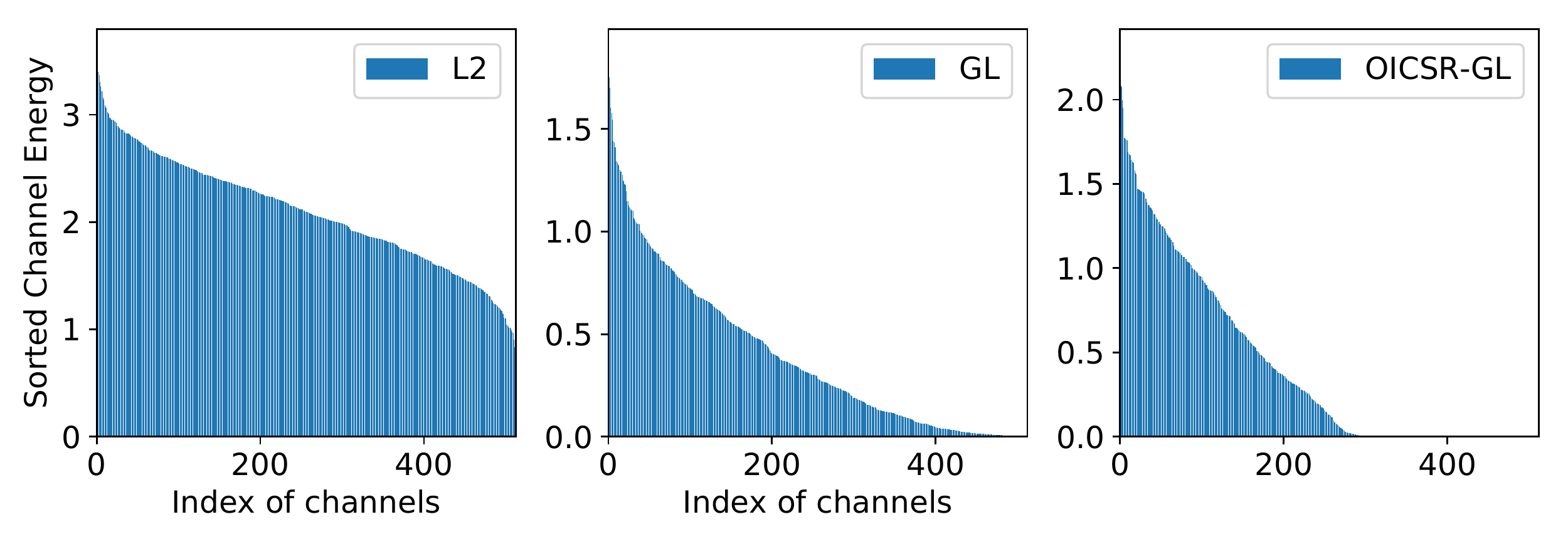}
\caption{The distribution of energy of out-in-channels (layer4.2.conv1 and layer4.2.conv2 in ResNet-56 on CIFAR-100 dataset) after training with L2, GL and OICSR-GL respectively. }
\label{energy}
\end{figure}
\begin{figure*} [t]
\centering
\subfloat [CifarNet on CIFAR-10 (84\% FLOPs pruned)] {\includegraphics[width=0.24\textwidth]{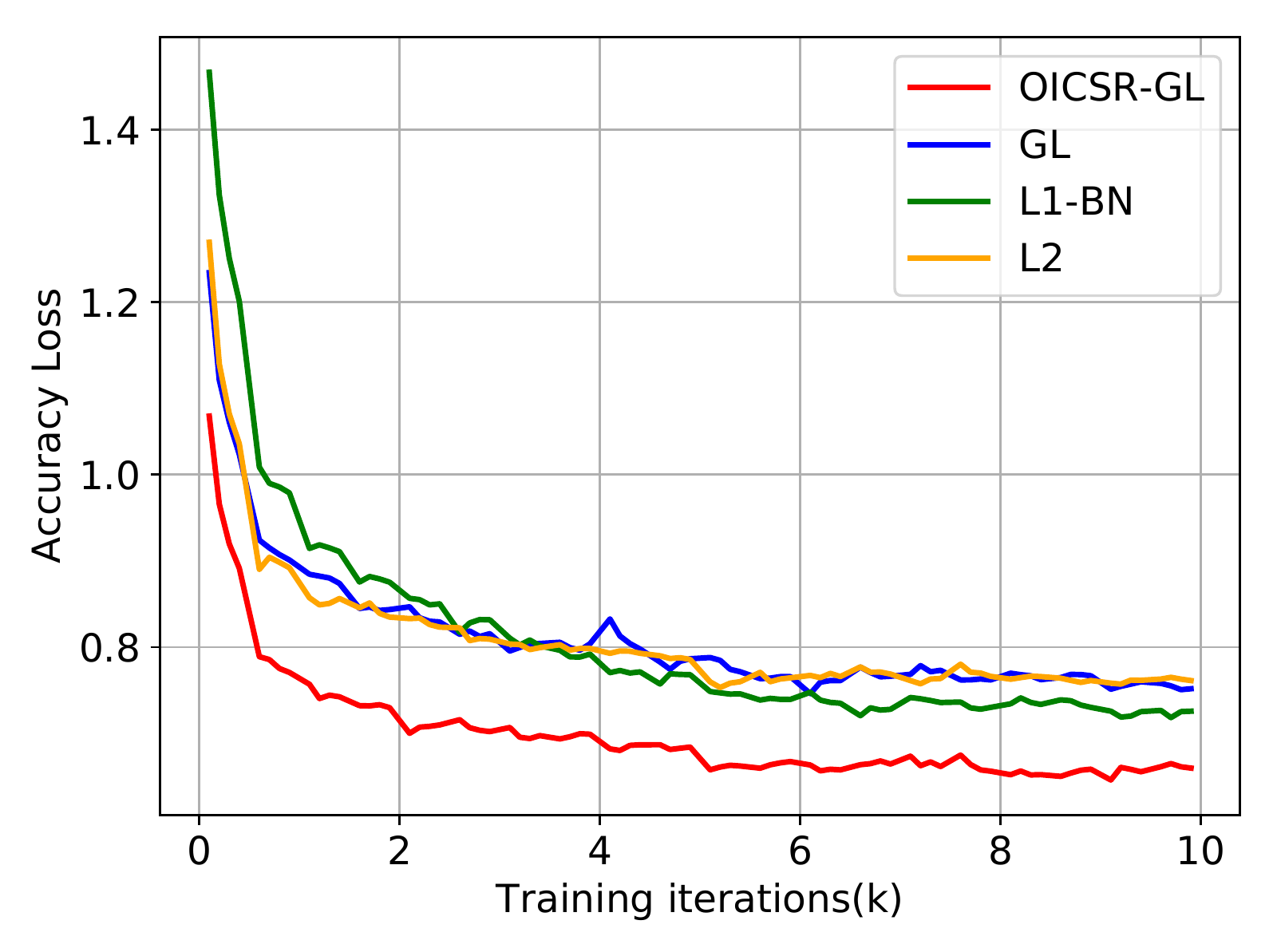}\label{cifarnet-cifar10}}
\hspace{1mm}
\subfloat [CifarNet on CIFAR-100 (74\% FLOPs pruned)] {\includegraphics[width=0.24\textwidth]{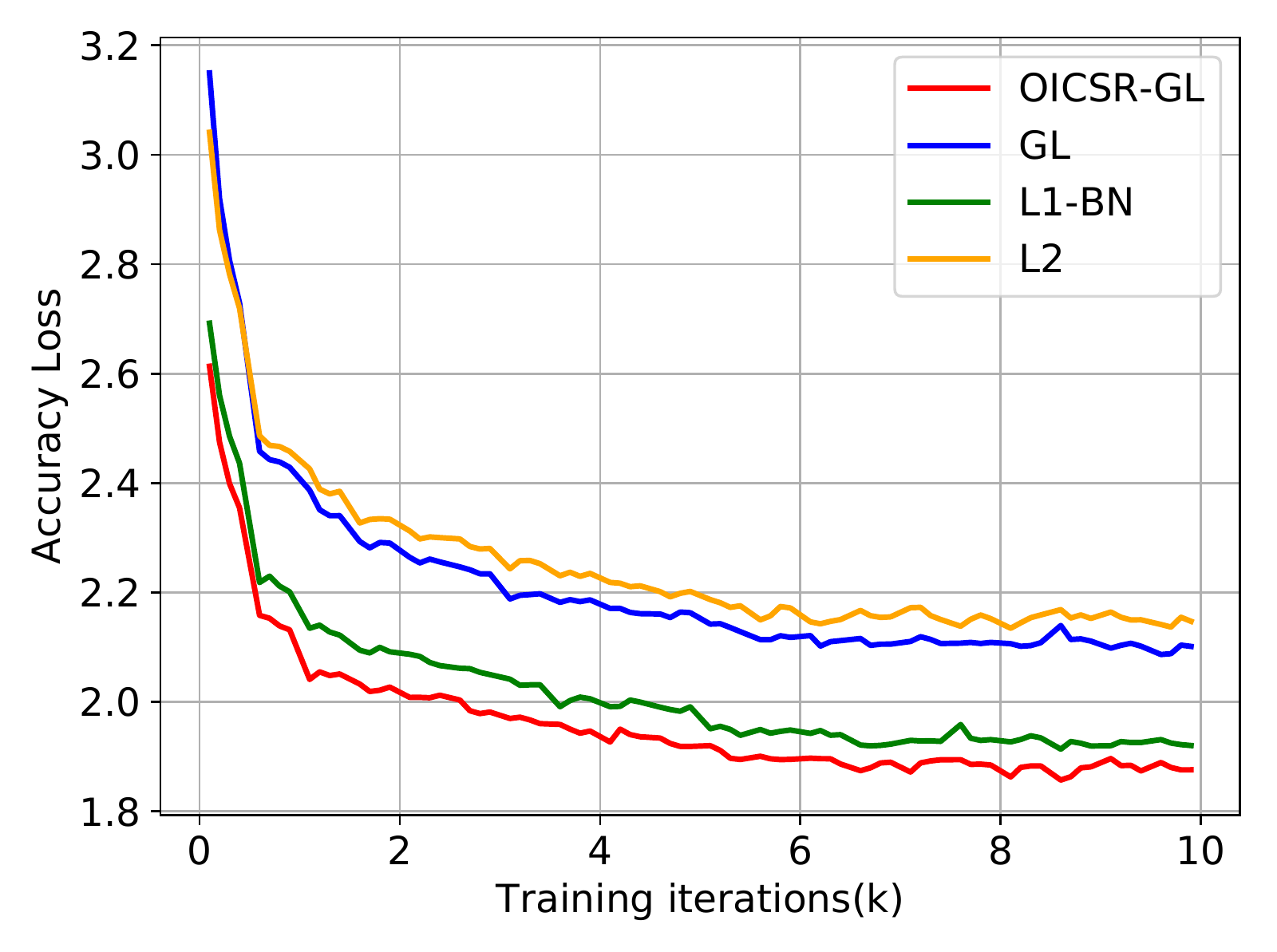}}
\hspace{1mm}
\subfloat [ResNet-18 on CIFAR-10 (94\% FLOPs pruned)] {\includegraphics[width=0.24\textwidth]{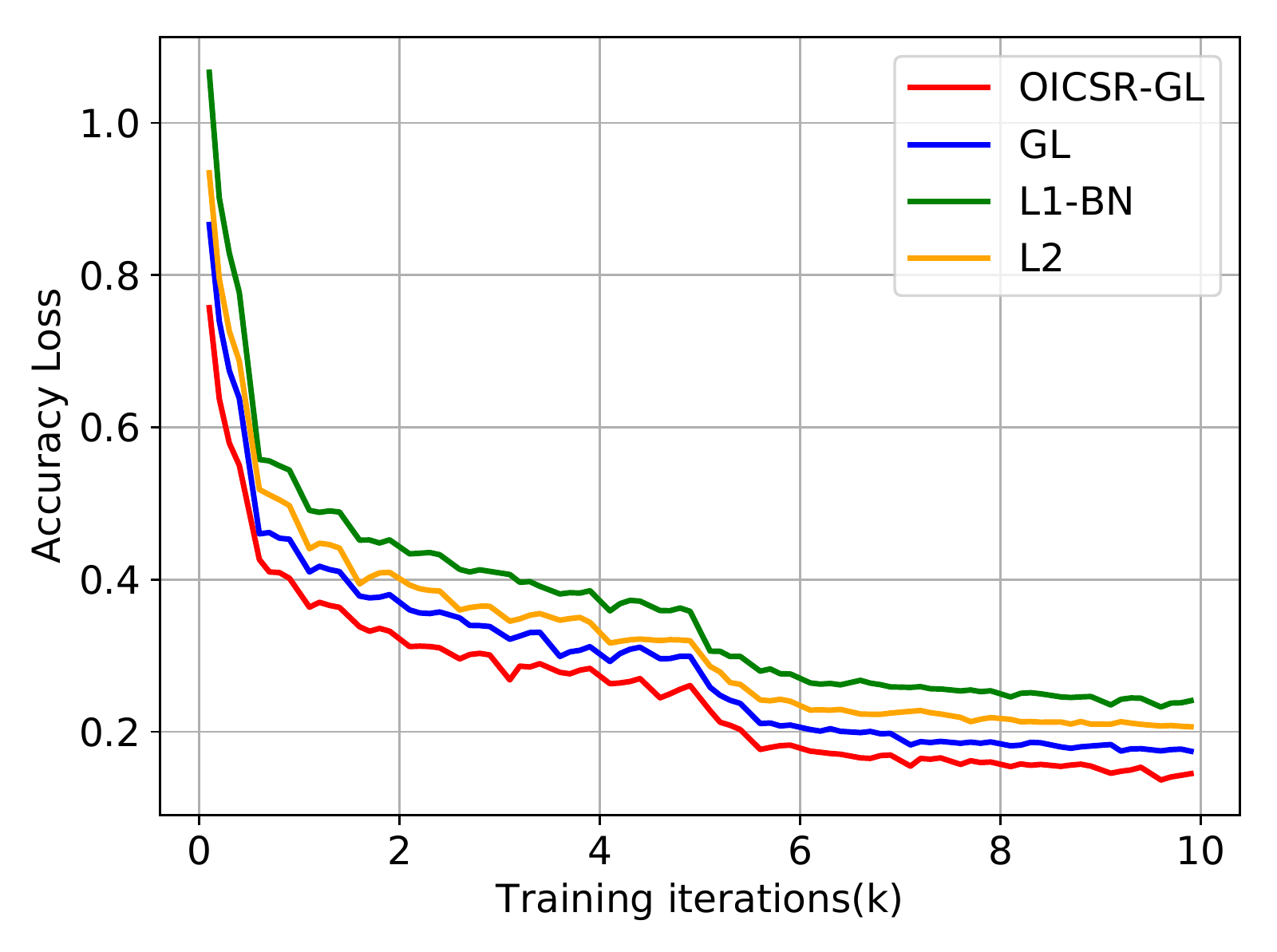}}
\hspace{1mm}
\subfloat [ResNet-56 on CIFAR-100 (97\% FLOPs pruned)] {\includegraphics[width=0.24\textwidth]{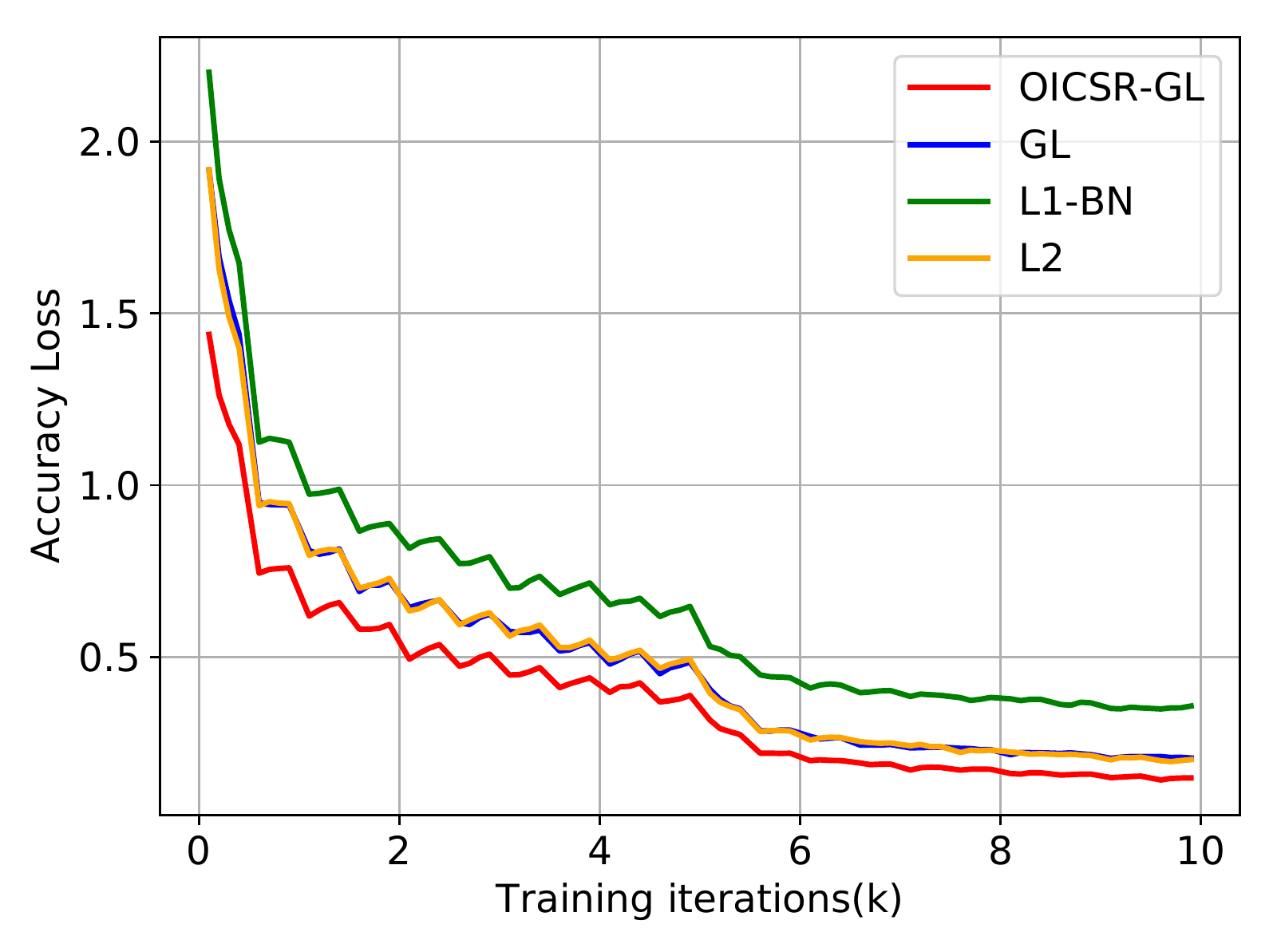}}
\hspace{1mm}
\caption{Accuracy loss of pruning/fine-tuning with OICSR and relevant baselines using CifarNet and ResNet on CIFAR-10 and CIFAR-100 datasets. Pruning/fine-tuning with OICSR-GL converges fastest with the lowest accuracy loss.}
\label{training_curve}
\end{figure*}
\begin{figure*} [t]
\centering
\subfloat [CifarNet on CIFAR-10 ] {\includegraphics[width=0.46\textwidth]{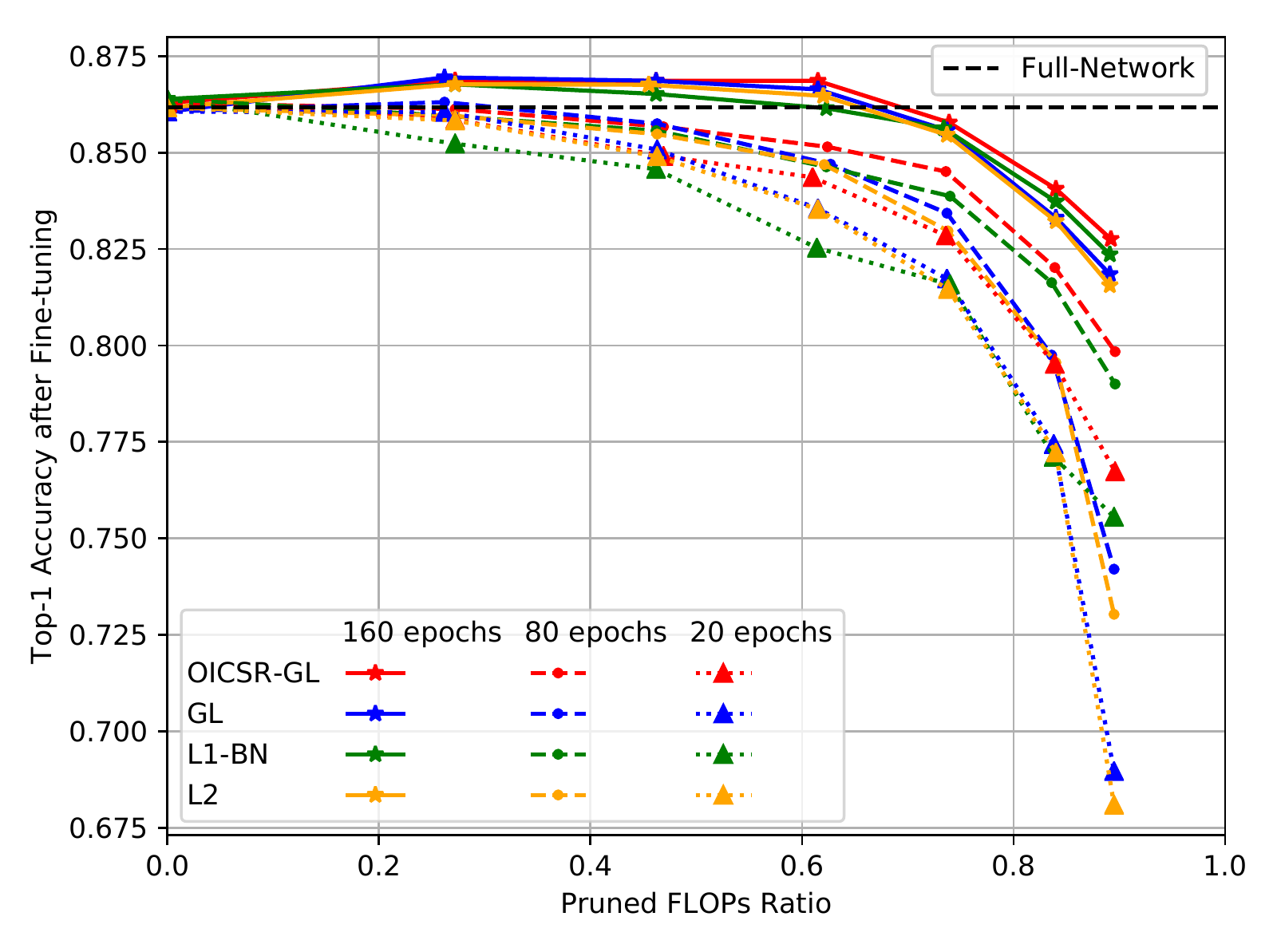}\label{cifarnet-cifar10}}
\hspace{5mm}
\subfloat [CifarNet on CIFAR-100 ] {\includegraphics[width=0.46\textwidth]{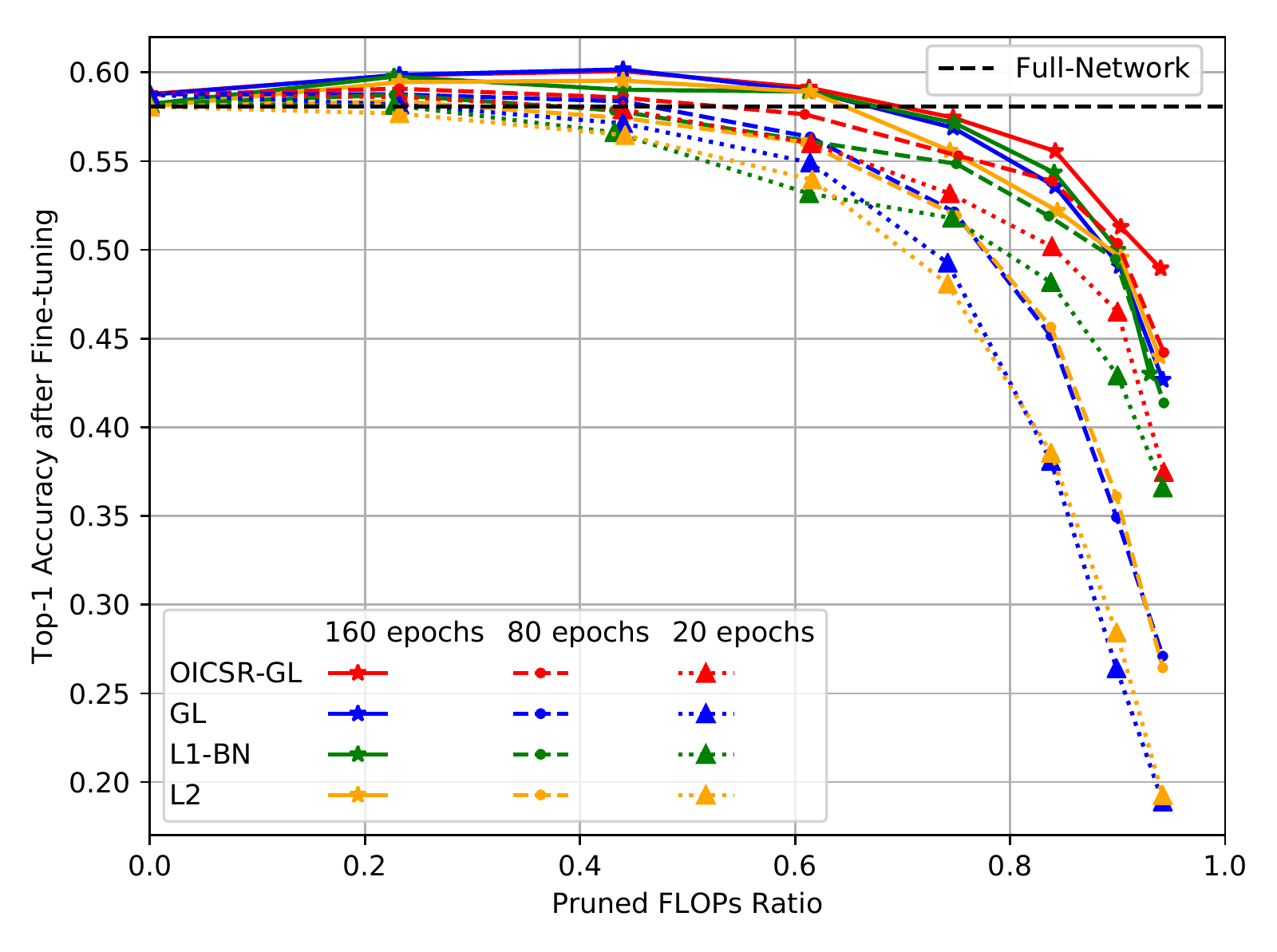}\label{cifarnet-cifar100}}
\caption{Comparison between OICSR-GL and relevant baselines of the trade-off between pruned FLOPs ratio and top-1 accuracy with various fine-tuning epochs. As expected, all the methods achieve higher accuracy with more fine-tuning epochs. With the same fine-tuning epochs, OICSR-GL consistently outperforms all the baselines under different pruned FLOPs ratio. }
\label{af_accuracy_cifarnet}
\end{figure*}
\subsection{Comparison with Non-structured Regularization and Seperated structured Regularization}
In this section, OICSR is compared with non-structured regularization and separated structured regularization from multiple aspects. The most classic structured regularization, Group Lasso~\cite{yuan2006model}, is chosen as the specific regularization term to demonstrate the effectiveness of OICSR. OICSR with Group Lasso (\textbf{OICSR-GL}) is used in all experiments. OICSR-GL is compared with three baselines: \textbf{(1) L2.} The network is only trained with non-structured regularization L2 regularization. The structure regularization
is not used.\textbf{(2) GL (Separated Group Lasso).} Group Lasso is separately applied on layers of the network as described in Eq.~\ref{equation 1}. \textbf{(3) L1-BN.} L1-BN~\cite{liu2017learning} is another form of separated structured regularization which applies L1-norm on the scaling factors of batch normalization layers. After training with L1-BN, we obtain a network in which the vector of scaling factors is sparse.  Scaling factors with small magnitudes and their corresponding channels are pruned. 

For all the above methods, the global greedy pruning algorithm (Algorithm~\ref{alg1}) is uniformly adopted to prune the redundant channels. All the experimental settings are the same except for the regularization and the corresponding criterion of channel importance. 
\subsubsection{Accuracy without Fine-tuning} \label{bf_accuracy}
We first validate whether OICSR-GL retains more important features and accuracy after channel pruning. The substantial remaining feature/accuracy works as a great initializer that leads to higher accuracy after fine-tuning~\cite{chin2018layer}. OICSR-GL is compared with relevant baselines by measuring top-1 accuracy (without fine-tuning) over pruned FLOPs. For fair comparison, results of all the methods are reported after channel pruning of the first iteration.

Fig.~\ref{accuracy-bf} shows the top-1 accuracy without fine-tuning of different classification tasks over pruned FLOPs, obtained by differentiating the structured regularization.  As expected, training and pruning with structured regularization GL, L1-BN and OICSR-GL reserve more prediction accuracy compared  with non-structured regularization L2 in most of cases. Separated Group Lasso is an efficient structured regularization which in general performs better than L1-BN and non-structured regularization L2. Specifically, L2, GL and OICSR-GL achieve similar performance (L1-BN performs worst) with CifarNet on CIFAR-10 dataset (Fig.~\ref{accuracy-bf}\subref{cifarnet-cifar10}) due to simplicity of both CIFAR-10 dataset and CifarNet architecture. Finally, OICSR-GL achieves the best trade-off between pruned FLOPs and prediction accuracy without fine-tuning owing to the fact that OICSR enforces channel-level sparsity on out-in-channels and prunes redundant out-in-channels based on statistical information computed from two consecutive layers. 

 For channel pruning, feature/energy in one out-in-channel is pruned/saved together. The distribution of energy of out-in-channels (Eq.~\ref{equation 8}) after training with different regularization are visualized (Fig.~\ref{energy}) to show the effect of OICSR. After training with OICSR-GL, energy distributes in less out-in-channels and more redundant channels are automatically selected. Here, OICSR-GL regularizes out-in-channels and transfers important features in much less out-in-channels compared with separated GL and non-structured regularization L2. Therefore, important features and accuracy can be maximally preserved by OICSR-GL after pruning the redundant out-in-channels.     
\subsubsection{Convergence Speed}
To further study advantages of OICSR-GL over baselines, we next analyze learning curves of different methods shown in Fig.~\ref{training_curve}. Interestingly, we find that L1-BN outperforms separated Group Lasso and L2 with CifarNet but performs worst with multi-branches architecture ResNet on CIFAR-10/100 datasets. For CifarNet on CIFAR-100 and ResNet-18 on CIFAR-10, separated Group Lasso converges faster and has lower accuracy loss compared with non-structured regularization L2.

OICSR-GL outperforms all the relevant baselines, which is reflected in three aspects. Firstly, OICSR-GL has less accuracy loss compared with baselines at fine-tuning iteration 0. This implies that OICSR-GL keeps higher accuracy after channel pruning, which agrees with the conclusion in section~\ref{bf_accuracy}. Secondly, OICSR-GL converges faster and achieves the same accuracy loss using much fewer fine-tuning iterations. Thirdly, OICSR-GL has lowest accuracy loss after fine-tuning. 
\begin{table}[]
	\centering
	
	\scalebox{0.92}{\subfloat[ResNet-18 on CIFAR-10 (Original Acc is 94.46\%)]{\begin{tabular}{p{1.50cm}cccc}
	\hline
	\diagbox{FLOPs}{Acc}{Regu} & L2 & L1-BN & GL & OICSR-GL \\
	\hline
	39.2\%$\downarrow$ 						  & 95.04\% & 94.95\% & 94.96\% & \bf{95.10\%}\\
	86.5\%$\downarrow$ 						  & 94.10\% & 93.77\% & 94.10\% & \bf{94.27\%}\\
	94.5\%$\downarrow$                  & 92.15\% & 91.02\% & 92.15\% & \bf{92.44\%}\\
	\hline
	\end{tabular}}}
	\hfil
	\scalebox{0.92}{\subfloat[DenseNet-89 on CIFAR-10 (Original Acc is 93.25\%)]{\begin{tabular}{p{1.30cm}cccc}
	\hline
	& L2 & L1-BN & GL & OICSR-GL \\
	\hline
	34.5\%$\downarrow$ 						  & 94.09\% & 94.09\% & 94.08\% & \bf{94.21\%}\\
	81.7\%$\downarrow$ 						  & 92.69\% & 92.37\% & 92.69\% & \bf{92.95\%}\\
	86.0\%$\downarrow$                  & 90.87\% & 91.21\% & 90.37\% & \bf{91.50\%}\\
	\hline
	\end{tabular}\label{table_densenet}}}
	\hfil
	\scalebox{0.92}{\subfloat[ResNet-56 on CIFAR-100 (Original Acc is 75.87\%)]{\begin{tabular}{p{1.30cm}cccc}
	\hline
	& L2 & L1-BN & GL & OICSR-GL \\
	\hline
	38.5\%$\downarrow$ 						  & 76.13\% & 75.28\% & 76.04\% & \bf{76.23\%}\\
	86.2\%$\downarrow$ 						  & 74.60\% & 75.13\% & 75.30\% & \bf{75.75\%}\\
	97.2\%$\downarrow$                  & 71.98\% & 72.36\% & 72.29\% & \bf{73.10\%}\\
	\hline
	\end{tabular}}}
	\hfil
	\scalebox{0.92}{\subfloat[PreActSeNet-18 on CIFAR-100 (Original Acc is 75.29\%)]{\begin{tabular}{p{1.30cm}cccc}
	\hline
	& L2 & L1-BN & GL & OICSR-GL \\
	\hline
	48.0\%$\downarrow$ 						  & 75.65\% & 75.76\% & 75.80\% & \bf{76.38\%}\\
	83.0\%$\downarrow$ 						  & 73.26\% & 72.66\% & 72.79\% & \bf{73.91\%}\\
	95.1\%$\downarrow$                  & 67.43\% & 65.52\% & 67.69\% & \bf{68.30\%}\\
	\hline
	\end{tabular}\label{table_senet}}}
	\hfil
	\scalebox{0.92}{\subfloat[AlexNet on ImageNet-1K (Original Acc is 56.98\%)]{\begin{tabular}{p{1.50cm}cccc}
	\hline
	& L2 & L1-BN & GL & OICSR-GL \\
	\hline
	23.4\%$\downarrow$ 						  & 57.62\% & ------ & 57.35\% & \bf{57.87\%}\\
	54.0\%$\downarrow$ 						  & 55.14\% & ------ & 55.02\% & \bf{56.83\%}\\
	68.3\%$\downarrow$                  & 52.55\% & ------ & 49.65\% & \bf{53.78\%}\\
	\hline
	\end{tabular}}}
	\hfil
	\scalebox{0.92}{\subfloat[ResNet-50 on ImageNet-1K (Original Acc is 76.31\%)]{\begin{tabular}{p{1.30cm}cccc}
	\hline
	& L2 & L1-BN & GL & OICSR-GL \\
	\hline
	37.3\%$\downarrow$ 						  & 76.39\% & 76.04\% & 76.23\% & \bf{76.53\%}\\
	44.4\%$\downarrow$ 						  & 76.03\% & 75.98\% & 76.02\% & \bf{76.30\%}\\
	50.0\%$\downarrow$                  & 75.80\% & 75.53\% & 75.76\% & \bf{75.95\%}\\
	\hline
	\end{tabular}}}
	\caption{Summary of the trade-off between top-1 accuray after fine-tuning and pruned FLOPs ratio with various CNNs on three benchmark datasets. [xx.x\%$\downarrow$] denotes the percentage of pruned FLOPs.}
	\label{af_accuracy_table}
\end{table}

\begin{table}[]
	\centering
	\scalebox{0.93}{\subfloat[ResNet-18 on CIFAR-10]{\begin{tabular}{lccc}
	\hline
	Methods & FLOPs$\downarrow$ & Params$\downarrow$ & Top-1 Acc$\downarrow$\\
	\hline
	Huang \etal~\cite{huang2018learning}		  & 35.30\% & --- & 1.00\%\\
	OICSR-GL 				  & \bf{39.20\%} & \bf{59.09\%} & \bf{-0.64\%}\\
	\hline
	Huang \etal~\cite{huang2018learning}	      & 76.00\% & --- & 2.90\%\\
	OICSR-GL                  & \bf{86.50\%} & \bf{90.89\%} & \bf{0.19\%}\\
	\hline
	\end{tabular}}}
	\hfil
	\scalebox{0.93}{\subfloat[AlexNet on ImageNet-1K]{\begin{tabular}{p{3.1cm}p{0.9cm}<{\centering}p{0.9cm}<{\centering}p{1.7cm}<{\centering}}
	\hline
	Methods & FLOPs$\downarrow$ & Params$\downarrow$ & Top-1 Acc$\downarrow$\\
	\hline
	FMP~\cite{molchanov2017pruning} (\cite{lin2018accelerating} impl.) & 37.62\% & --- & 1.87\%\\
	GDP~\cite{lin2018accelerating} & 52.30\% & --- & 0.77\%\\
	NISP~\cite{Yu_2018_CVPR} 		  & 53.70\% & 2.91\% & 0.54\%\\
	OICSR-GL          & \bf{54.00\%} & \bf{3.06\%} & \bf{0.15\%}\\
	\hline
	\end{tabular}}}
	\hfil
	\scalebox{0.93}{\subfloat[ResNet-50 on ImageNet-1K]{\begin{tabular}{p{2.2cm}ccc}
	\hline
	Methods & FLOPs$\downarrow$ & Top-1 Acc$\downarrow$ & Top-5 Acc$\downarrow$\\
	\hline
	LcP~\cite{chin2018layer}         & 25.00\% & -0.09\% & -0.19\%\\
	NISP~\cite{Yu_2018_CVPR}		 & 27.31\% & 0.21\% & ---\\
	SSS~\cite{huang2017data}     & 31.08\% & 1.94\% & 0.95\%\\
	ThiNet~\cite{luo2017thinet:}     & 36.79\% & 0.84\% & 0.47\%\\
	OICSR-GL                         & \bf{37.30\%} & \bf{-0.22\%} & \bf{-0.16\%}\\
	\hline
	He \etal~\cite{he2018soft}	     & 41.80\% & 1.54\% & 0.81\%\\
	GDP~\cite{lin2018accelerating}   & 42.00\% & 2.52\% & 1.25\%\\
	LcP~\cite{chin2018layer}         & 42.00\% & 0.85\% & 0.26\%\\
	NISP~\cite{Yu_2018_CVPR} 	     & 44.41\% & 0.89\% & ---\\
	OICSR-GL                         & \bf{44.43\%} & \bf{0.01\%} & \bf{0.08\%}\\
	\hline
	He \etal.~\cite{he2017channel}   & 50.00\% & --- & 1.40\%\\
	LcP~\cite{chin2018layer}         & 50.00\% & 0.96\% & 0.42\%\\
	OICSR-GL                         & 50.00\% & \bf{0.37\%} & \bf{0.34\%}\\
	\hline
	\end{tabular}}}
	\caption{Comparison with existing methods. FLOPs$\downarrow$ and Params$\downarrow$ denote the reduction of FLOPs and parameters. Top-k Acc$\downarrow$\ denotes the decline of top-k accuracy and a negative value indicates an improvement of model accuracy.}
	\label{comparison_other_method}
\end{table}
\subsubsection{Accuracy after Fine-tuning} \label{af_accuracy}
OICSR-GL is also evaluated over relevant baselines in terms of prediction accuracy after fine-tuning. The results with various fine-tuning epochs for CifarNet on CIFAR-10/100 datasets are reported in Fig~\ref{af_accuracy_cifarnet}. As expected, under the same pruned FLOPs ratio, all the methods achieve higher accuracy with more fine-tuning epochs. The loss feature/accuracy induced by channel pruning can be well retrieved by fine-tuning. The fewer fine-tuning epochs, the more obvious the advantage of OICSR-GL is. With the same fine-tuning epochs, OICSR-GL consistently performs better than all the baselines under different pruned FLOPs ratios. Moreover, OICSR-GL fine-tuned with 20 epochs achieves higher accuracy compared with baselines fine-tuned with 80 epochs in certain sparsity ranges. 

Results of the other network architectures are reported in Table~\ref{af_accuracy_table}. To maintain channel-level sparsity during pruning, the networks on  CIFAR-10/100 dataset and ImageNet-1K are fine-tuned with 160 epochs and 60 epochs respectively for each channel pruning iteration. Shown in Table~\ref{af_accuracy_table}, the superiority of OICSR-GL gradually emerges with the increase of pruned FLOPs ratio. Pruning channels and fine-tuning with OICSR-GL lead to higher generalization accuracy. For ResNet-18 on CIFAR-10 dataset, OICSR-GL obtains 0.64\% accuracy improvement using 39.2\% less FLOPs and achieves 7.4$\times$ FLOPs reduction with only 0.19\% top-1 accuracy drop. OICSR-GL improves 1.09\% accuracy while using 48.0\% less FLOPs for PreActSeNet-18 on CIFAR-100 dataset. OICSR speeds up ResNet-56 by 7.2$\times$ with only 0.12\% top-1 accuracy loss on CIFAR-100 dataset. For AlexNet and ResNet-50 on ImageNet dataset, OICSR gains 0.89\% and 0.32\% accuracy improvement while using  23.4\% and 37.3\% less FLOPs; OICSR-GL also achieves 2.2$\times$ and 2.0$\times$ speedup with only 0.19\% and 0.36\% top-1 accuracy decline respectively. Moreover, as shown in Table~\ref{af_accuracy_table}\subref{table_densenet} and Table~\ref{af_accuracy_table}\subref{table_senet}, OICSR-GL outperforms relevant baselines on both popular networks DenseNet-89 and PreActSeNet-18.
\subsection{Comparison with Other Methods}
We compare our method with other state-of-the-art channel pruning techniques (not regularization-based) using AlexNet and ResNet on CIFAR-10 and ImageNet-1K datasets. As shown in Table~\ref{comparison_other_method}, for ResNet-18 on CIFAR-10, OICSR-GL significantly reduces more parameters (90.89\%) and FLOPs(86.50\% vs. 35.30\%~\cite{huang2018learning}), while achieving less accuracy decline (0.19\% vs. 1.00\%~\cite{huang2018learning}). 

Our method also shows superior performance on ImageNet-1K dataset. For AlexNet, with less accuracy loss, OICSR-GL prunes more FLOPs(54.00\%) compared with FMP~\cite{molchanov2017pruning} (37.62\%); OICSR reduces similar FLOPs but achieves much less accuracy loss (0.15\%) compared with GDP~\cite{lin2018accelerating} (0.77\%) and NISP~\cite{Yu_2018_CVPR} (0.54\%). We are the first to exploit regularization-based channel pruning methods for very deep residual network ResNet-50 on ImageNet-1K dataset. Under various pruned FLOPs ratios, our method consistently achieves state-of-the-art result compared with prior arts~\cite{chin2018layer,he2018soft,he2017channel,huang2017data,lin2018accelerating,luo2017thinet:,molchanov2017pruning,Yu_2018_CVPR}, which strongly aligns with our pervious analysis and observation.
\section{Conclusion}
Current deep neural networks are effective with high inference costs. In this paper, we propose a novel structured regularization form, namely OICSR, which takes account correlations between successive layers to learn more compact CNNs. OICSR regularizes out-in-channels and measures channel importance based on statistical information of two consecutive layers. To minimize accuracy loss caused by incorrect channel pruning, we investigate a global greedy pruning algorithm to select and remove redundant out-in-channel in an iterative way. As a result, important features and accuracy are greatly preserved by OICSR after channel pruning. Experiments demonstrated the superiority of OICSR against non-structured regularization and separated structured regularization. Furthermore, our method achieves better results compared with existing state-of-the-art channel pruning techniques. 

\footnotesize{\noindent\textbf{Acknowledgements.}\quad This work was supported in part by the National Natural Science Foundation of China under Grant 61671079, Grant 61771068, and Grant 61471063, and in part by the Beijing Municipal Natural Science Foundation under Grant 4182041.}
{\small
\bibliographystyle{ieee}
\bibliography{egbib}
}

\end{document}